\documentclass{bmcart}
\usepackage{graphicx}
\usepackage{amsthm,amsmath}
\usepackage{amsfonts}
\usepackage{url}
\usepackage{subcaption} 
\usepackage{blindtext}
\usepackage{tikz}
\usetikzlibrary{spy}
\usepackage[utf8]{inputenc} 

\startlocaldefs
\endlocaldefs

\begin{document}

\begin{frontmatter}

\begin{fmbox}
\dochead{Research}


\title{Deep image prior inpainting of ancient frescoes in the Mediterranean Alpine arc}


\author[
  addressref={aff1,aff2},                   
  corref={aff1},                       
  email={fabio.merizzi@unibo.it}   
]{\inits{F.M.}\fnm{Fabio} \snm{Merizzi}}
\author[
  addressref={aff4},
  email={perrine.saillard@etu.univ-cotedazur.fr}
]{\inits{P.S.}\fnm{Perrine} \snm{Saillard}}
\author[
  addressref={aff4},
  email={Oceane.acquier@univ-cotedazur.fr}
]{\inits{P.S.}\fnm{Oceane} \snm{Acquier}}
\author[
  addressref={aff3},
  email={elena.morotti4@unibo.it}
]{\inits{E.M.}\fnm{Elena} \snm{Morotti}}
\author[
  addressref={aff1},
  email={elena.loli@unibo.it}
]{\inits{E.L.P.}\fnm{Elena} \snm{Loli Piccolomini}}
\author[
  addressref={aff2},
  email={calatroni@i3s.unice.fr}
]{\inits{L.C.}\fnm{Luca} \snm{Calatroni}}
\author[
  addressref={aff4},
  email={Rosa-Maria.Dessi@unice.fr}
]{\inits{R.M.D.}\fnm{Rosa Maria} \snm{Dessì}}


\address[id=aff1]{
  \orgdiv{Department of Computer Science and Engineering (DISI)},     
  \street{via Mura Anteo Zamboni 7},                       
  \postcode{40126}  
  \orgname{University of Bologna},          
  \city{Bologna},                              
  \cny{Italy}                                    
}
\address[id=aff2]{%
  \orgdiv{Laboratoire d'Informatique, Signaux et Syst\`{e}mes de Sophia-Antipolis (i3S)},
  \orgname{CNRS, Inria, Universit\'e Côte d'Azur, Les Algorithmes - b\^at. Euclide B},
  \street{2000, route des Lucioles},
  \postcode{06903}
  \city{Sophia-Antipolis},
  \cny{France}
}
\address[id=aff3]{%
  \orgdiv{Department of Political and Social sciences},
  \orgname{University of Bologna},
  \street{Str. Maggiore 45},
  \postcode{40125}
  \city{Bologna},
  \cny{Italy}
}
\address[id=aff4]{%
  \orgdiv{Cultures et Environnements Pr\'ehistoire, Antiquit\'e, Moyen \^Age},
  \orgname{Universit\'e C\^ote d'Azur, P\^ole Universitaire Saint Jean d’Ang\'ely},
  \street{ 24, avenue des Diables Bleus},
  \postcode{06300}
  \city{Nice},
  \cny{France}
}



\end{fmbox}

\begin{abstractbox}

\begin{abstract} 
The unprecedented success of image reconstruction approaches based on deep neural networks has revolutionised both the processing and the analysis paradigms in several applied disciplines. In the field of digital humanities, the task of digital reconstruction of ancient frescoes is particularly challenging due to the scarce amount of available training data caused by ageing, wear, tear and retouching over time. To overcome these difficulties, we consider the Deep Image Prior (DIP) inpainting approach which computes appropriate reconstructions by relying on the progressive updating of an untrained convolutional neural network so as to match the reliable piece of information in the image at hand while promoting regularisation elsewhere. In comparison with state-of-the-art approaches (based on variational/PDEs and patch-based methods), DIP-based inpainting reduces artefacts and better adapts to contextual/non-local information, thus providing a valuable and effective tool for art historians. As a case study, we apply such approach to reconstruct missing image contents in a dataset of highly damaged digital images of medieval paintings located into several chapels in the Mediterranean Alpine Arc and provide a detailed description on how visible and invisible (e.g., infrared) information can be integrated for identifying and reconstructing damaged image regions.


\end{abstract}




\end{abstractbox}
%

\end{frontmatter}


\section{Introduction}  \label{sec:intro}

The synergy between art history, mathematical image analysis and artificial intelligence (AI) is a stimulating meeting point between disciplines to favour the development of new science and to complement historical studies in art and art history. These new tools and methods lead to an emerging approach in the comprehension of medieval images as living objects, see, e.g., \cite{dessi_spectres}. 
In this work we focus on the digital reconstruction of wall paintings of medieval chapels located in the south of the Alpine arc. The wall paintings in this area were produced mainly between the second half of the 15th century and the early 16th century \cite{dessi_dataset}. We are interested in particular in the wall paintings signed or attributed to the painters Giovanni Baleison and Tommaso and Matteo Biazaci. They were active in the last quarter of the 15th century in current France and Italy. Their peculiarity is the frequent use of texts in their painted images. As part of several restoration campaigns and/or more specific modifications linked to the shift of perception and reception of the images depicted in the murals, such paintings have been subject to modifications in later times. 
Furthermore, the effect of the environment and/or the intentional erasure and vandalism caused the disappearance of several imaging data crucial for the understanding of some images and painted texts. \\
In order to digitally restore the missing/lost image elements made indecipherable by such processes, digital reconstruction approaches and among them, image inpainting \cite{Bertalmio2000}, can be applied, see \cite{fornasier2007restoration, Baatz2008, calatroni2018unveiling} for previous applications in digital humanities contexts. Given the lack of information, the restoration of the original version of the degraded image under consideration is impossible (inpainting is indeed an ill-posed problem lacking uniqueness) so the objectives of inpainting in this context are rather concerned to the reconstruction of a coherent visual experience to the observer, which may help the comprehension and interpretation of damaged images in historic studies. 
Moreover, a careful analysis of the output images may shed light on whether the observed corruptions are involuntary or intentional, thus generally favouring a better understanding of the overall artistic process. 
By combining inpainting with multi-spectral techniques, interesting piece of information can be unveiled, such as the stratification of murals and the evolution of images over time. 
A further aim of our digital reconstructions is to determine both the dates and the authors of each image layer which, compared to major artworks, are still debated. From a historical viewpoint, our objective is to grasp the causes at the roots of transformations that may be aesthetic, religious, or ideological. In this way, we think this interdisciplinary project between art history, mathematical image processing, and AI, can allow us to chronicle the life of the paintings and better understand their impact and evolution in past societies. 
The reconstruction of digital images of frescoes characterized by large occlusions with irregular shapes is a very challenging task. 
A large variety of the inpainting approaches proposed in the literature rely either on the expert choice of the reconstruction model by the user \cite{Bugeau2010,schoenlieb_2015} or on the use of large training sets of data \cite{ballester2022analysis}, which both limit their practical use in the field of digital humanities.
We consider an unsupervised neural approach for the 
 digital inpainting of images of highly damaged frescoes. Our method belong to the class of so-called {\it Deep Image Prior} algorithms \cite{Ulyanov_2020}.
Compared to supervised approaches relying on large data sets of examples, the proposed approach is fully unsupervised and performs reconstruction based only on the observation of the damaged image and on the detection of the region to be filled in. 
We detail in this work how such existing approach can be applied to the challenging task of digital reconstruction of highly damaged frescoes and highlight the modifications performed both in the neural architecture and in the DIP loss function to improve both performance and stability. Our setting is proved to be effective in comparison to state of the art approaches and validated on both simulated and real data including, e.g., the restoration of textual characters and the use of infrared data for the study of the transformation/retouching process the artworks have been subject to. 
This manuscript is organized in the following manner: In Section \ref{sec:dataset} the image dataset used for our study is described and enriched with information on the artistic/historical context. In Section \ref{sec:inpainting} a comprehensive discussion on state-of-art inpainting methods is given, covering both handcrafted and data-driven approaches. 
In Section \ref{sec:inp_DIP}, we introduce the DIP approach and our proposal.
In Section \ref{sec:experSetup}, the overall pipeline of our approach is described, spanning from the initial treatment and analysis performed on the given image to inpaint till the final inpainted result. Several numerical results are reported in Section \ref{sec:results} where comparisons between inpainting approaches and combined techniques making use of both visible and invisible (infrared) data are combined, thus showing the potential of the proposed approach to the study of imaging data in digital humanities.
At last, we draw our conclusions in Section \ref{sec:conclusions}.

\section{Dataset description and challenges}  
\label{sec:dataset}

\begin{figure}
	\begin{subfigure}{\textwidth}
			\centering
			\includegraphics[height=45mm]{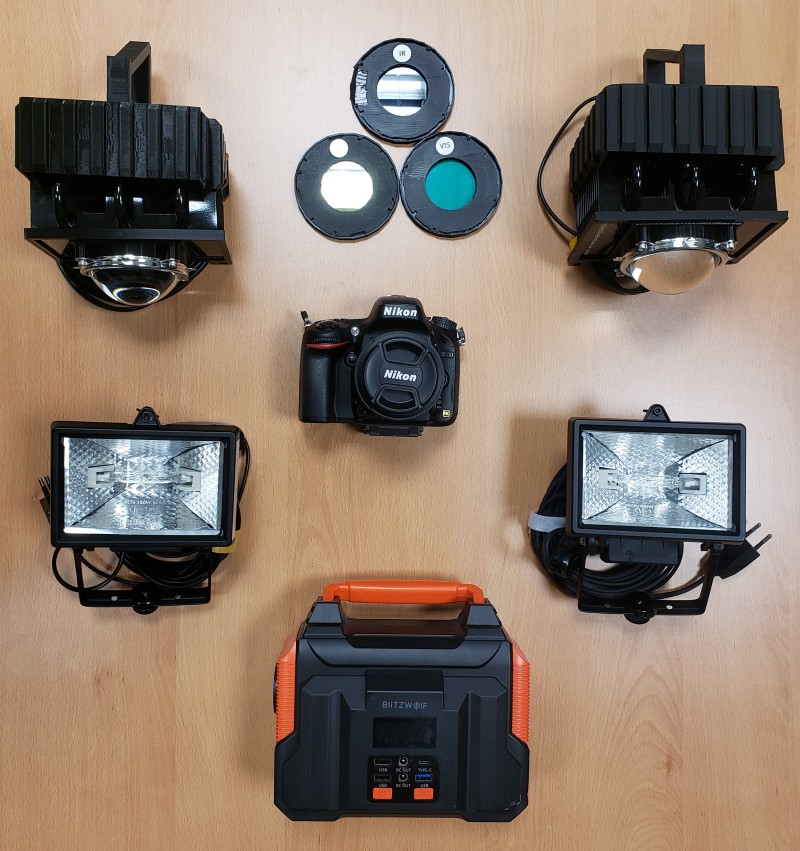}
	\includegraphics[height=45mm]{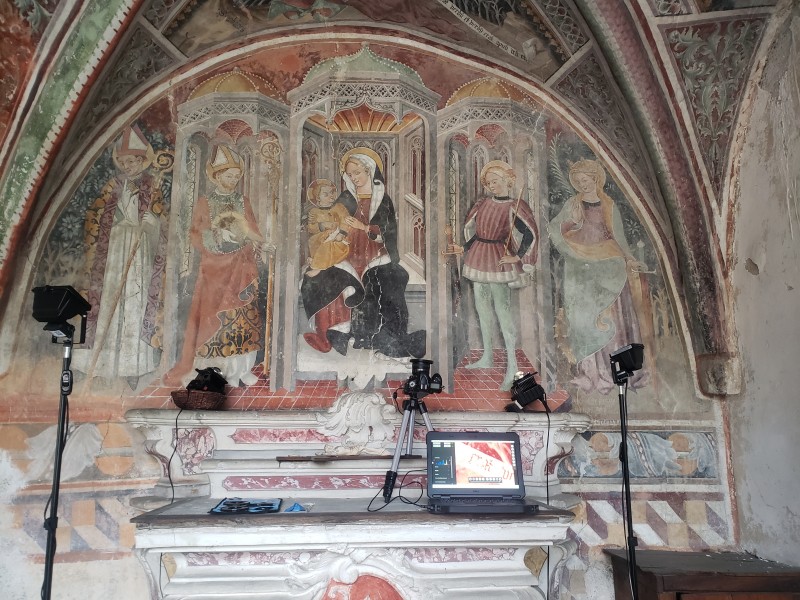}
	\caption{Cameras, filters and acquisition setting}
	\label{fig:setting}
        \begin{subfigure}{\textwidth}
			\centering
			\includegraphics[width=0.9\linewidth]{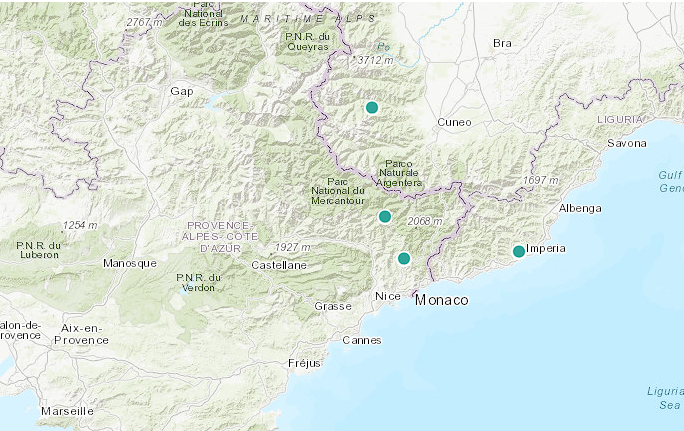}
			\caption{Locations of the four chapels, along the Alpine arc between France and Italy.}
			\label{fig:locations}
	\end{subfigure}
\end{subfigure}
\caption{Locations, devices and experimental setup for data acquisition.}
\label{fig:setup}
\end{figure}

The image dataset used in this project has been collected in the online database PA'INT \cite{dessi_dataset} (CEPAM, UCA, FR) which has been collected as part of the PhD thesis of O. Acquier \cite{dessi_tesiAcquier}. The database is composed by a large collection of digital images of late medieval wall paintings representing visual scenes and epigraphic items in religious buildings of the south of the Alpine arc. 
In total, 269 painted monuments have been geolocated of which 75 have been the object of several image acquisition campaigns. As a result, $~$2600 pictures have been collected and indexed to various details such as the name of the painter(s) (when known), the date(s) of completion as well as a visual descriptions. A total number of 1172 inscriptions have been analysed in \cite{dessi_tesiAcquier}. Note that currently PA'INT is in the process of being expanded with images in the infrared and ultraviolet spectral range, which will be analysed and integrated by means of AI tools in a later work. 
The images in the dataset have been acquired by a modified Nikon D610\footnote{Our digital camera has been modified by EOS FOR ASRTO.} \cite{dessi_galli}, in which a filter that blocks ultraviolet and infrared (IR) has been removed, with the Nikon AF-S NIKKOR 50mm f/1.8G lens. In order to limit the light reception to the desired spectral range, some light filters were used corresponding to a wavelength of 380-780 nm for the visible spectrum  and 780-1100 nm for the infrared spectrum. Flashes BOWENS GEMINI 1500 pro as well as lighter and less bulky halogen lamps from CHSOS \cite{dessi_TP} were used, see Figure \ref{fig:setting}. For the infrared emissions, halogen lamps are placed at approximately 45$^{\circ}$ of the studied painted surfaces, which were also captured in the visible range for comparisons/data-integration, see Figure \ref{fig:setup}. The interest of IR acquisitions is that they can reveal retouches and underwritings if the overpainter layer is IR-transparent and the underpaintings are not. For some references on the use of scientific imaging in digital humanities, we refer to \cite{Boust}.\\

As a case study,  we analysed incomplete and retouched images of wall paintings acquired in four chapels: the chapel {\it Sainte-Claire}\footnote{Also called chapel of Saint Sébastien because of the representation of the saint.}  in Venanson, France, the sanctuary {\it Nostra Signora delle Grazie} in Imperia, Italy, the chapel \emph{Notre Dame de Bon Coeur} in Luc\'eram, France and the chapel {\it San Sebastiano} in Celle di Macra, Italy. See Figure \ref{fig:locations} for their geolocalizations. 

\begin{figure}
\centering
\includegraphics[width=0.9\textwidth]{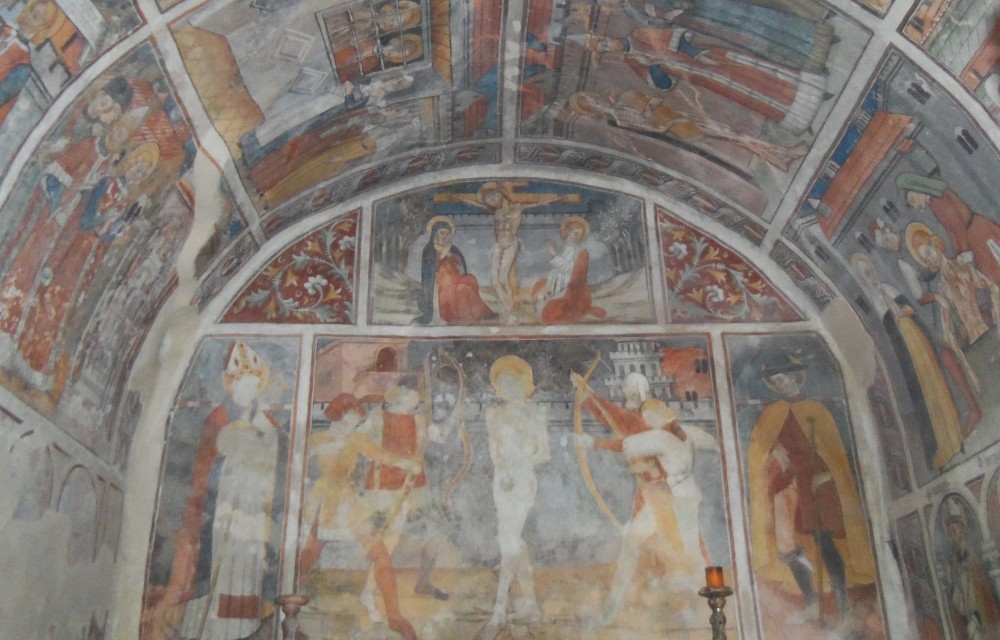}
\caption{Martyrdom of S. Sébastien in Venanson.}
\label{fig:venanson}
\end{figure}

The decoration of the \emph{Sainte Claire} chapel was painted by Giovanni Baleison in 1481. The Venanson community had this chapel constructed, and the decorations were commissioned by Guillaume Cobin, as indicated in the signature (Figure \ref{fig:venanson_text}). It is best known as the Saint Sébastien chapel because a large portion of the wall paintings is dedicated to the life of saint Sebastian, and his martyrdom is depicted in the chevet of the chapel, see Figure \ref{fig:venanson}. Unlike the frescoes in Celle di Macra and Montegrazie, the chapel walls do not depict Hell. However, they still feature, like Nostra Signora delle Grazie, the theme of \emph{cavalcade} of vices, a popular motif in the Alps during that period.

The sanctuary of \emph{Nostra Signora delle Grazie} has undergone at least four decoration campaigns since the late 15th century. In this paper, we will focus on the frescoes painted by the Biazaci brothers in 1483 (Figure \ref{fig:DIPonIR_christ}) and by Pietro Guido da Ranzo between 1524 and 1540 (Figure \ref{fig:DIPonIR_arc}). The decorations were overpainted during the 18th century and were rediscovered during restoration campaigns throughout the 20th century. The images presented in this paper illustrate the virtues of \emph{charitas} and \emph{sobrietas} as painted by Tommaso and Matteo Biazaci  and details from Pietro Guido's Mocking of Christ, respectively.
The wall paintings from the chapel Notre Dame de Bon Coeur are attributed to either Giovanni Baleison or the Master of Luc\'eram. The decoration was executed between 1480 and 1485.



\begin{figure}[h!]
 \centering
\includegraphics[height=0.3\textwidth]{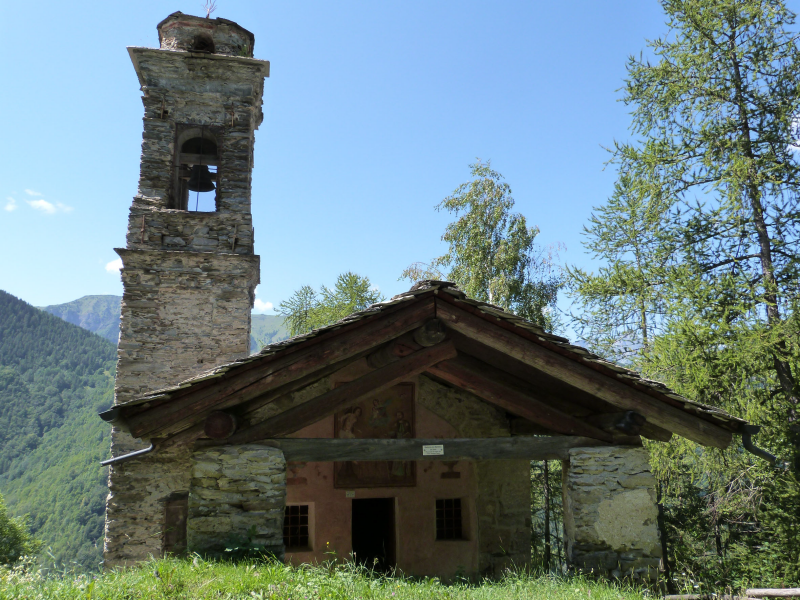}
\includegraphics[height=0.3\textwidth]{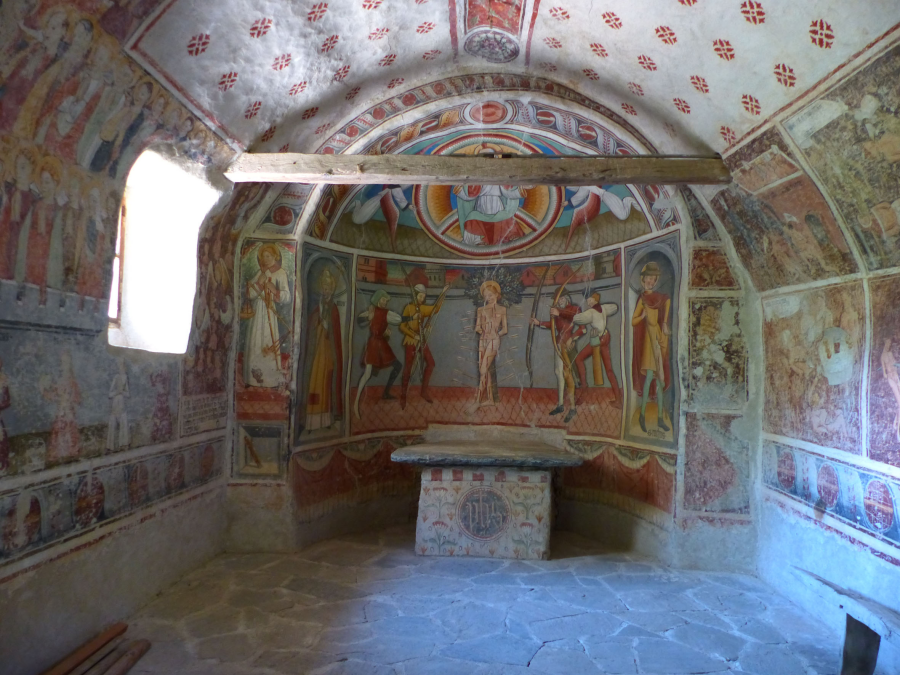} \\
\includegraphics[width=0.9\textwidth]{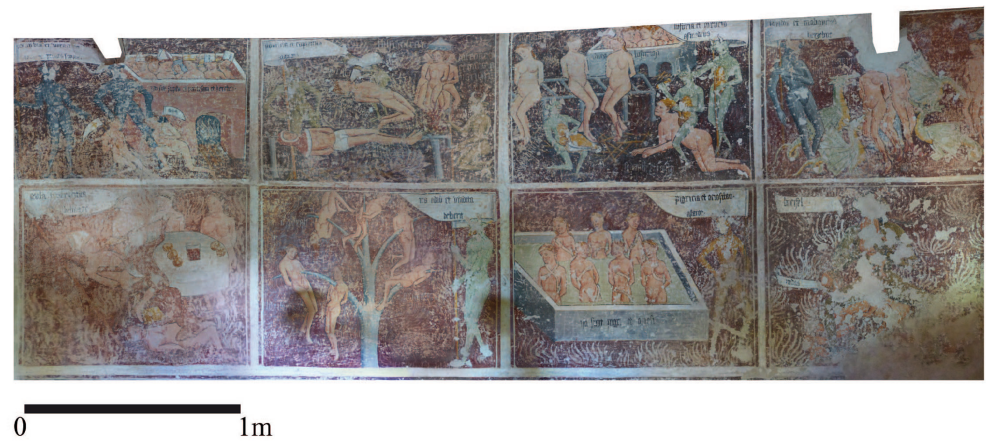}
\caption{The chapel of San Sebastiano  in Cella di Macra, Italy. 
}
\label{fig:san_sebastiano2}
\end{figure}

\begin{figure}[h!]
\begin{subfigure}[b]{0.45\textwidth}
  \centering
  \includegraphics[width=\linewidth]{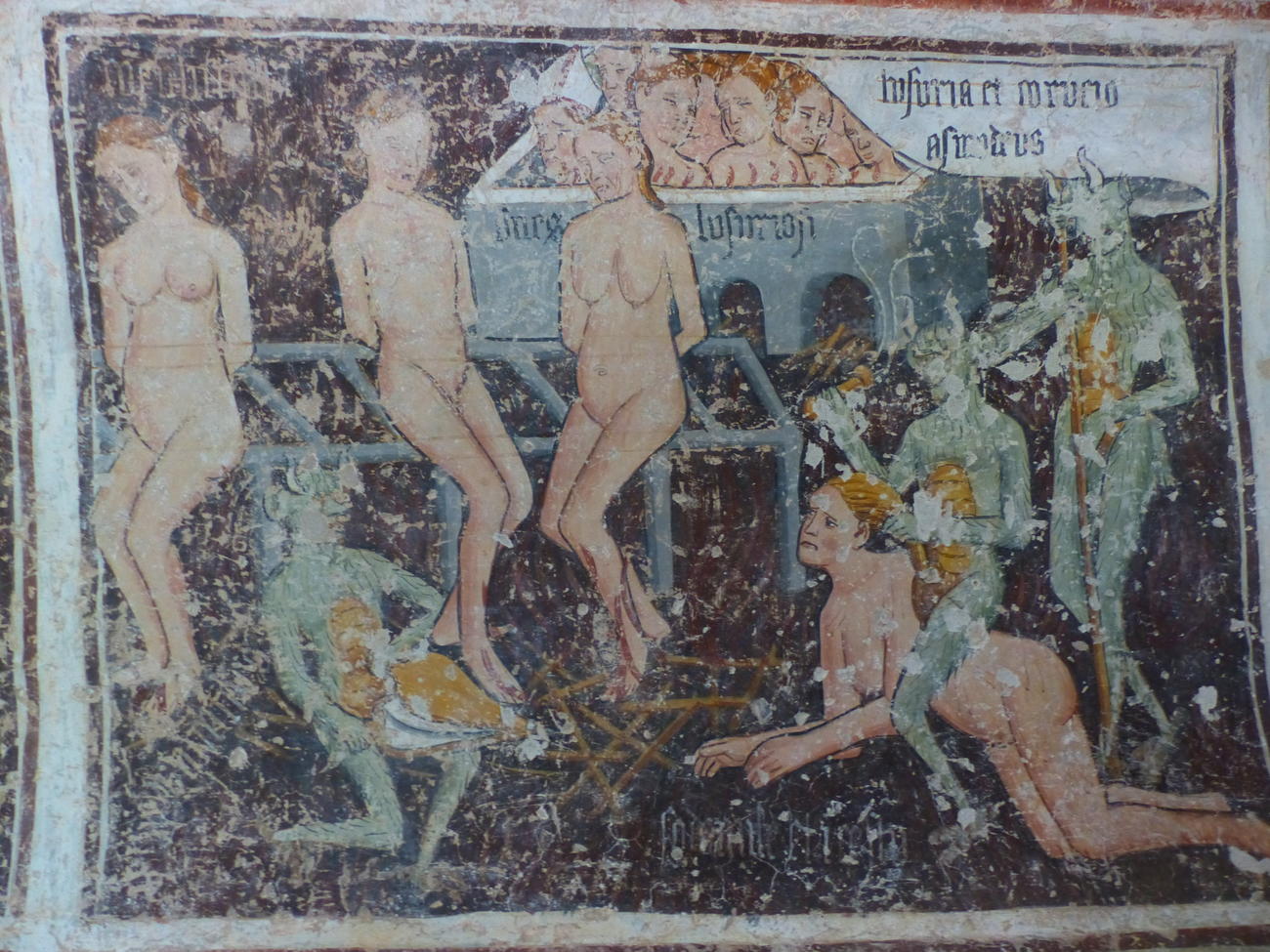}  
  \caption{\emph{Lusuria}}
 	\label{fig:sub-luxuria}
\end{subfigure}
\begin{subfigure}[b]{0.45\textwidth}
	\centering
	\includegraphics[width=\linewidth]{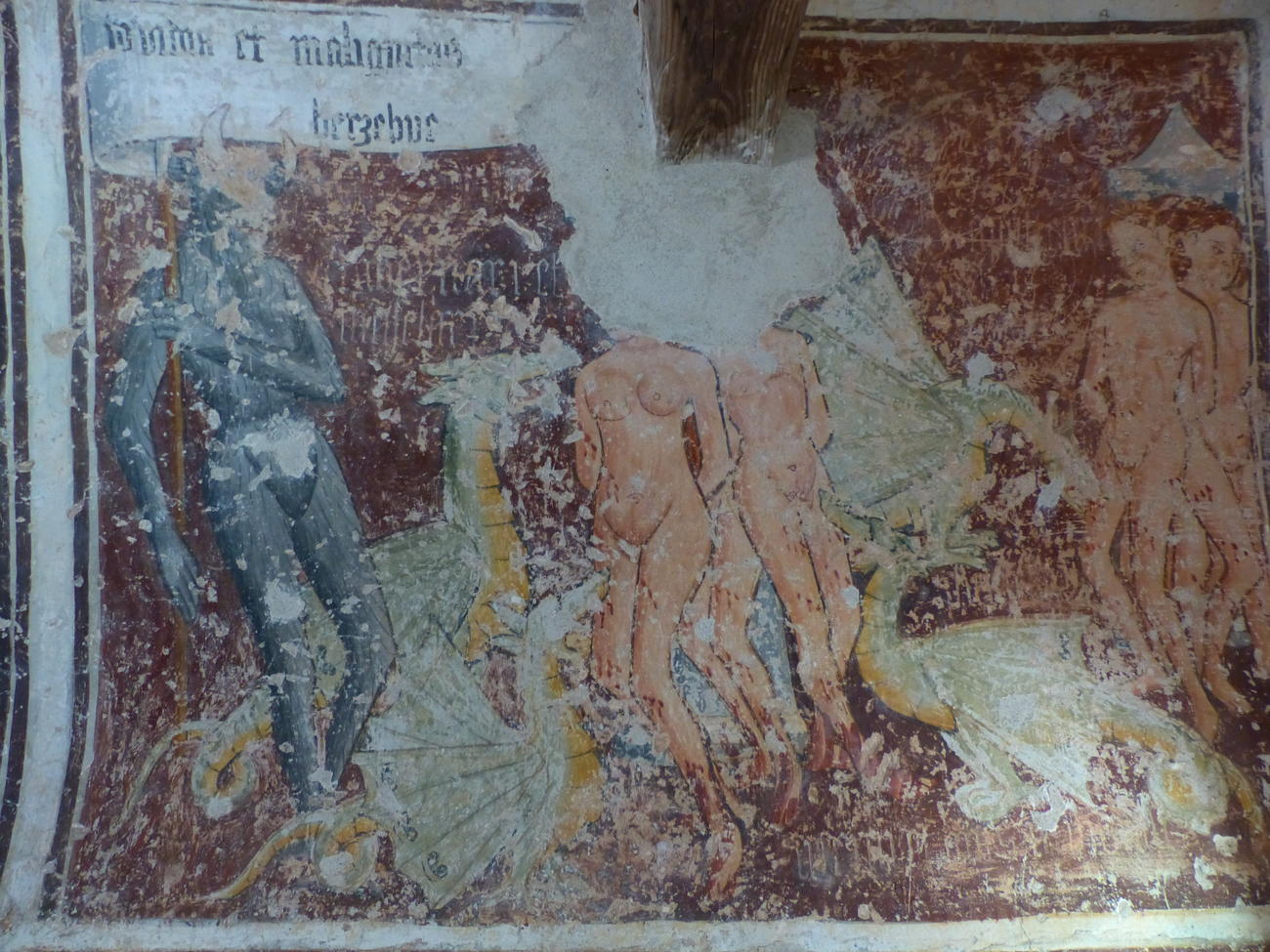}  
	\caption{\emph{Invidia}}
  \label{fig:sub-invidia}
\end{subfigure}
\caption{Two selected scenes from the chapel of San Sebastiano in Cella di Macra, from Figure \ref{fig:san_sebastiano2}.}
\label{fig:san_sebastiano}
\end{figure}

Figure \ref{fig:san_sebastiano2} shows the chapel of San Sebastiano in Celle di Macra and the representation of Hell painted therein by Giovanni Baleison in 1484. The fresco is divided into eight parts, among which seven are dedicated to a particular capital sin, while the last one is Lucifer’s den. In this work, we will focus in particular on the images of \emph{Lusuria} and \emph{Invidia}, see Figure \ref{fig:san_sebastiano}. The scene represented in \emph{Lusuria}, Figure \ref{fig:sub-luxuria}, is ruled by the demon Asmodeus. Its circle welcomes souls prone to lust and carnal pleasures in their earth life. In this scene, green and yellow demons are torturing sinners: a demon is whipping a woman while pulling her hair. Three sinners are sitting on a grill fed by a demon, while a group of men and women are burning inside a building. \emph{Invidia}, see Figure \ref{fig:sub-invidia}, constitutes the fourth infernal pit, ruled by the blue demon Belzebub. The pit hosts sinners culpable of envy and malignancy. The demon is accompanied by four green and yellow dragons which are painted in the action of lacerating sinners. The damned souls are divided into two groups, each composed by three persons tied up to a spike. 
Due to the extensive deterioration of these paintings, responsible for making numerous painted texts present in the background not understandable and prone to possible misinterpretations. A digital reconstruction procedure is expected to facilitate the understanding of the written text and, overall, of the painted scene.

\section{State-of-art methods for image inpainting}  \label{sec:inpainting}

The problem of image inpainting consists of filling in missing or damaged parts of an image (representing, e.g., a fresco) using a source of prior information.

In mathematical terms, given a colour image $\tilde{x}$ defined on an image domain $\Omega =\left\{ (i,j): i=1,\ldots, m, j=1,\ldots, n \right\}$ of size $m\times n$ having an occluded region $D\subset \Omega$, the problem is defined in terms of a masking operator $m\in \left\{0,1\right\}^{m\times n}$ acting point-wise as follows : 
\begin{eqnarray}
m_{i,j} =  
    \begin{cases}
      1 & \text{if $\bar{x}_{i,j} \in \Omega\setminus  D $ }\\
      0 & \text{if $\bar{x}_{i,j} \in D$ }.
    \end{cases}
\label{eq:mask}
\end{eqnarray}
By definition, the mask $m$ is thus nothing but the characteristic function of the set $\Omega\setminus D$ and identifies the reliable (i.e., unoccluded) pixels in the observed image.

Most of the classical approaches employed over the last three decades rely on the use of mathematical approaches favouring the transfer of the available image content  within the region to be filled in by means of diffusion/transport processes and/or by copy-paste procedures of appropriate patches. 

Often, their design requires a certain modelling expertise aimed at choosing which type of diffusion (linear VS. non-linear, for instance) is preferred for the image at hand. We will refer to this class of approaches as \emph{hand-crafted} approaches, meaning by that name the fact that they are designed by an expert user. As their numerical implementation often relies on the use of iterative algorithms, these approaches have been also called \emph{sequential} algorithms in the recent literature \cite{elharrouss2020image}. We provide a review of these methods and of their main features in Section \ref{ssec:inpainting_MB}.

More recent techniques rely on the shared idea of filling in the incomplete image regions by novel image content generated by neural networks trained on large image datasets \cite{ballester2022analysis}. Due to the prominent role played by the data for this class of approaches, we will refer to them as \emph{data-driven} approaches and describe their main features in Section \ref{ssec:inp_DD}.

In the following paragraphs we review the main available literature on both approaches, with a particular attention to their application to their use in the field of cultural heritage.


\subsection{Inpainting by hand-crafted approaches}  \label{ssec:inpainting_MB}


Hand-crafted methods for digital image inpainting have been actively proposed since the early 2000s. 
The most famous approaches are based on local diffusion techniques, which can fill the missing regions by diffusing image information locally, from the known image portions into the adjacent damaged ones, at the pixel level, see, e.g. \cite{schoenlieb_2015,Bugeau2010} for reviews.
These approaches model the problem in a variational form where the inpainted image $\hat{x}$ solves:
\begin{equation}
\hat{x}\in \text{argmin}_x  ~ \lambda||m \odot (x - \bar{x})||^2 + R(x), 
\label{eq:var_energy}
\end{equation}
where the data term forces $x$ to stay close to the data  $\bar{x}$ on $\Omega\setminus D$  and $R(\cdot)$ is a regularisation term favoring the propagation of contents within $D$. The effect of regularization against data fidelity is weighted by $\lambda>0$. In the data term, the symbol $\odot$  stands for the Hadamard element-wise product. Partial Differential Equation (PDE) approaches stem from \eqref{eq:var_energy} by considering the corresponding Euler-Lagrange equations, possibly embedded within an artificial evolution towards the minimizer(s) of the corresponding functional.\\
A popular instance of \eqref{eq:var_energy} proposed in \cite{Chan2001} consists in choosing a regularization term $R(\cdot)$ favouring piece-wise constant reconstructions via non-linear diffusion. This can be done by choosing $R(x) = TV(x)$, the Total Variation (TV) regularization functional which acts on images as:
\begin{equation}
TV(x) =  \sum_{c\in\left\{R, G, B \right\}}\sum_{i=1}^{m-1} \sum_{j=1}^{n-1} \sqrt{ (x^c_{i+1,j}-x^c_{i,j})^2 + (x^c_{i,j+1}-x^c_{i,j})^2 },
\label{eq:TVprior}
\end{equation}
where $x^c_{i,j}$ denotes the intensity value of the $c\in\left\{R, G, B \right\}$ channel of the image at pixel $(i,j)\in \Omega$.\\
More complex choices can be made at a variational level such as, e .g., higher-order regularization (see, e.g., \cite{Papafitsoros2014}). On the other hand, from a PDE viewpoint, advanced approaches making use of Navier-Stokes models propagating colour information by means of complex diffusive fluid dynamics laws have been considered in \cite{Caselles1998, Bertalmio2000, bertalmio2001navier, Telea2004, Sapiro}. Other approaches involved the use of transport and curvature-driven approaches \cite{Ballester2001,chan2001nontexture, Masnou1998}.


Being based on the discretization of differential operators, the hand-crafted approaches described above favour \emph{local} regularization. As a consequence, they are particularly suited to reconstruct only small occluded regions such as scratches, text, or similar. In the context of heritage science, they have been employed  for restoring ancient frescoes in works such as \cite{Bertalmio2000,fornasier2007restoration,Baatz2008} showing effective performance.

On the other hand, such techniques fail in reconstructing large occluded regions and in the retrieval of more complex image content such as texture. To overcome such limitation, \emph{non-local} inpainting approaches have been proposed in a variety of papers (see, e.g. \cite{Criminisi2004,Aujol2010,Arias2011}) to propagate image information using patches. 
In more detail, the main idea consists of comparing patches from the known image regions in terms of a suitable similarity metric which can further take into account rigid transformations and/or patch rescaling. The popularised PatchMatch approach \cite{Barnes2009} is based on this principle, with the further advantage of computing correspondence probabilities for each patch and thus weighting the contribution coming from different locations appropriately. Improved versions of PatchMatch have been proposed, e.g., in \cite{Newson2014,ipol.2017.189} where such averaging is performed in a non-local manner. Compared to local approaches, patch-based inpainting methods show remarkable performance and, where properly tuned, good reconstruction of both geometric and textured contents. Nonetheless, due to their intrinsic non-convexity, they are often initialization dependent and are sensitive to the choice of hyperparameters such as, e.g., the patch size. In the context of art restoration, in \cite{calatroni2018unveiling} a combination of a local (as initialization) and non-local (as the main inpainting process) procedure was used for the digital restoration of severely damaged illuminated manuscripts.

 An interesting comparison between local/non-local sequential approaches for the inpainting of digital images of artworks has been conducted in \cite{Oncu2012}. Interestingly, the authors therein noted that while manual restoration still seems to lead to the best results, reconstructions obtained by model-based approaches appear often misleading for expert evaluation, while as good as a manual reconstruction for naïve eyes.

The choice of the most appropriate hand-crafted model (in particular, of the most appropriate term $R(\cdot)$ favouring inpainting within $D$) often requires some technical modelling expertise. This limits the use of this class of approaches in practice, as an optimal choice of such term typically requires the understanding of advanced concepts in linear/non-linear diffusion and smooth/non-smooth optimisation which are highly non-standard for practitioners.

\subsection{Inpainting by data-driven approaches}  \label{ssec:inp_DD}

Data-driven approaches for image inpainting offer an alternative strategy to the conventional methods of modeling image regularity through predefined energy functionals. Instead, these methods leverage an extensive array of training data and employ neural techniques to estimate mappings from occluded input images to inpainted images.  Due to their better deep encoding capabilities, neural approaches are indeed not limited to the modeling of the sole geometric/texture regularities in an image, but they further capture the presence of local/non-local patterns and the semantic meaning of image contents.

An exhaustive review of learning-based approaches for image inpainting is presented in \cite{ballester2022analysis}. Upon prior knowledge of the inpainting region, i.e. of the mask operator in \eqref{eq:var_energy}, data-driven inpainting approaches based on convolutional networks have been designed in \cite{Khler2014MaskSpecificIW,Pathak2016} and improved in some recent works such as \cite{Liu2018,Wang2018}, with the intent to adapt the convolutional operations only to those points providing relevant information.



The performance of data-driven inpainting dramatically improved after the introduction of the generative adversarial network (GAN) architectures in \cite{NIPS2014_5ca3e9b1}. GANs aim to minimize the distance between ground truth images and reconstructed images not in a point-wise manner, but, rather, in a distributional sense, through the use of two competing networks, the former able to discriminate between ground truth data and samples generated by the latter. Whenever a large number of examples is available, GANs and, more in general, generative neural approaches, are very effective for inpainting, see, e.g. \cite{Pathak2016,Iizuka2017,Liu2019,Liu_2021_CVPR, Lahiri_2020_CVPR, HEDJAZI2021106789}. Improved approaches perform inpainting by working, rather than at an image level, at the level of feature space, by first reconstructing the  geometric content and finally adding finer textures, see for instance \cite{Ren2019,Xiong2019}.



More recently, Denoising Diffusion Probabilistic Models (DDPM) \cite{ho2020denoising} have emerged with comparable and possibly overall greater inpainting performance than GANs. DDPMs can achieve optimal results in generative tasks without the impairment typical of GAN models, such as adversarial learning instabilities and high computational cost \cite{goodfellow2015explaining}. A recent effort in inpainting with diffusion models reported impressive results  \cite{lugmayr2022repaint} by conditioning the reverse diffusion process with mask information. 
Other recent examples of neural data-driven inpainting techniques based, e.g., on diffusion models include \cite{chen2023crackdiffusion, Wang_2023_CVPR,Li_2022_CVPR,Suvorov_abc}. 

Despite their excellent performance, data-driven approaches have scarcely been used to perform digital inpainting tasks. Some examples are, e.g., \cite{Wang2021,Lv2022,Deng2023} where (generative) learning approaches are employed. In order to generate suitable image contents, these approaches require the availability (or the synthetic generation) of large datasets of relevant and high-quality data and occlusion type for training. This constitutes indeed a major limitation in the reconstruction of highly-damaged frescoes painted by local authors for which, therefore, very little training data is available.\\
Generally speaking, the use of data-driven approaches to solve the problem of digital inpainting is often limited due, essentially, to:
\begin{itemize}
    \item The scarce availability of reference data to be used for training;
    \item The bias induced by non relevant data during inpainting.
\end{itemize}

\section{Deep Image Prior  inpainting}  \label{sec:inp_DIP}

To overcome the limitations of the approaches described before, we will consider in the following a tailored approach, popularised under the name of \emph{Deep Image Prior} (DIP) in \cite{Ulyanov_2020}. This approach  combines the interpretability of hand-crafted regularisation models with the power of data-driven methods. It employs a neural procedure to inpaint the image and, in comparison to classical learning schemes, makes use of the sole observed image as a training example. 

This technique pioneers the use of low-level image statistics extracted from an image by the network structure itself, hence DIP allows to obtain an accurate inpainted image without a training set, exploiting an expressive untrained architecture on just one degraded image. In other words, DIP enables the use of a neural technique in our specific inpainting application.  

\begin{figure}
\centering
\includegraphics[width=0.9\textwidth]{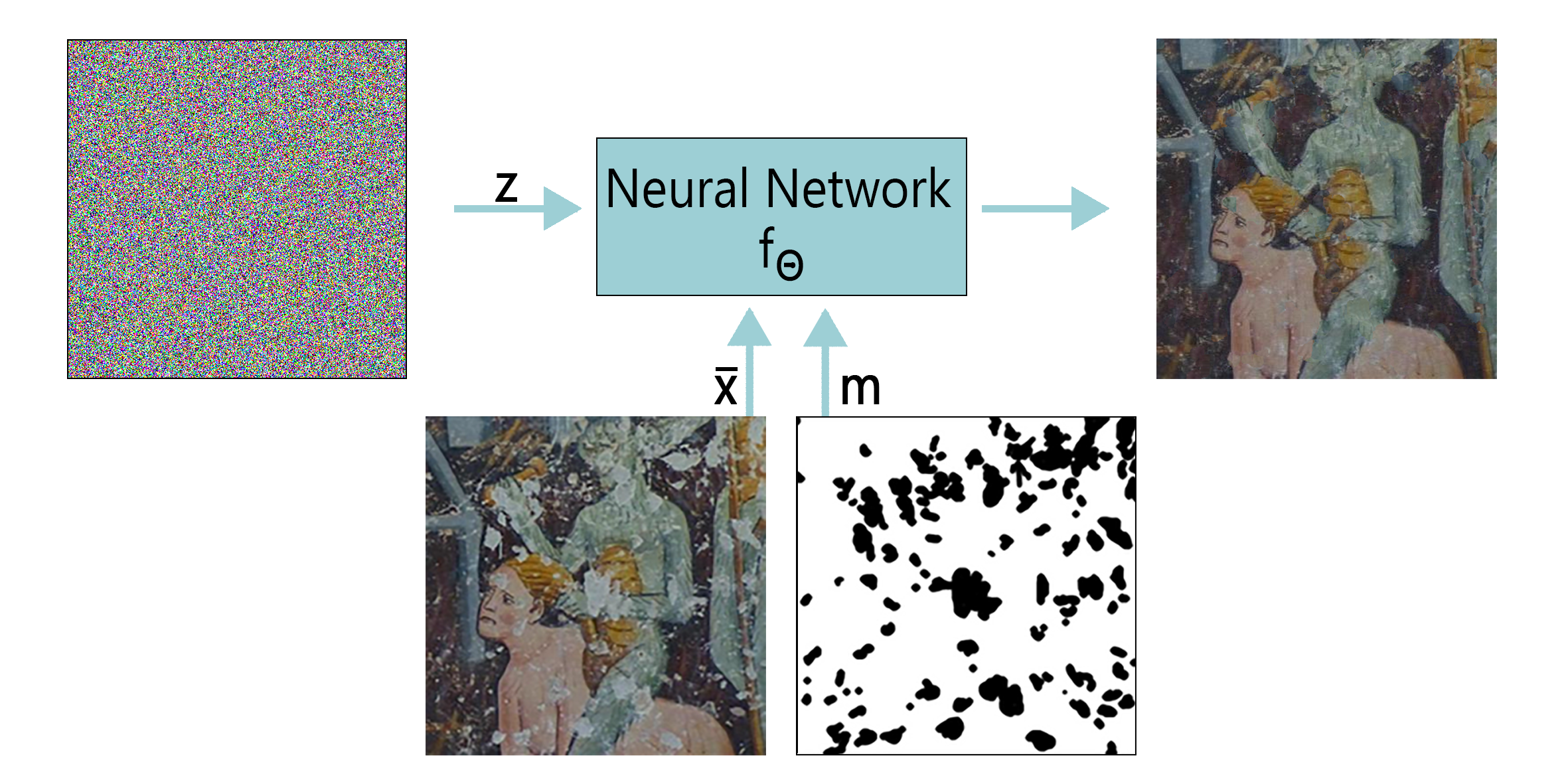}
\caption{DIP inpainting methodology. The network is fed random noise $z$, original image $\bar{x}$, and binary mask $m$, to produce as output the inpainted image.}
\label{fig:dip}
\end{figure}

In Figure \ref{fig:dip}, we graphically represent how DIP works for the inpainting problem at hand. In particular, we show that the neural network takes as input an image $z$,  randomly sampled from a uniform distribution with a variable number of channels, and it also considers the damaged image $\bar{x}$ and its corresponding mask $m$, then it gives as output the restored image.  
Formally, the DIP approach computes the vector of neural network parameters $\hat{\Theta}$ by solving the minimisation problem:
\begin{equation}
\hat{\Theta}\in\text{argmin}_\Theta ~ ||m \odot (f_{\Theta}(z) - \bar{x})||^2,
\label{eq:deep_prior_network}
\end{equation}
where $f_{\Theta}(\cdot)$ is a neural network with parameters $\Theta$. 
By solving \eqref{eq:deep_prior_network}, the parameters $\hat{\Theta}$ generate an output image $\hat{x} = f_{\hat{\Theta}}(z)$ matching at best $\bar{x}$ outside $D$ and filling contents in $\Omega\setminus D$.
Numerically, this problem can be solved by standard iterative optimisation algorithms such as gradient descent with back-propagation. Being \eqref{eq:deep_prior_network} a non-convex optimisation problem, different initialisations for $\Theta$ may lead to different results. Note that DIP implicitly enforces regularisation through the network structure, unlike traditional methods, but the early stopping of iterations is necessary to avoid overfitting.\\
Clearly, the training procedure \eqref{eq:deep_prior_network} depends on the given image $\bar{x}$ to be inpainted.
In case several images are to be restored, the weights must be recomputed for each degraded image, independently. 
As a consequence, the DIP computational cost is more similar to the one of model-based methods than to data-driven approaches, where the parameters are computed only once using large exemplar sets with a very expensive training phase.

\subsection{DIP architecture and regularisation}  \label{ssec:networks}

\begin{figure}
\centering
\includegraphics[width=0.9\textwidth]{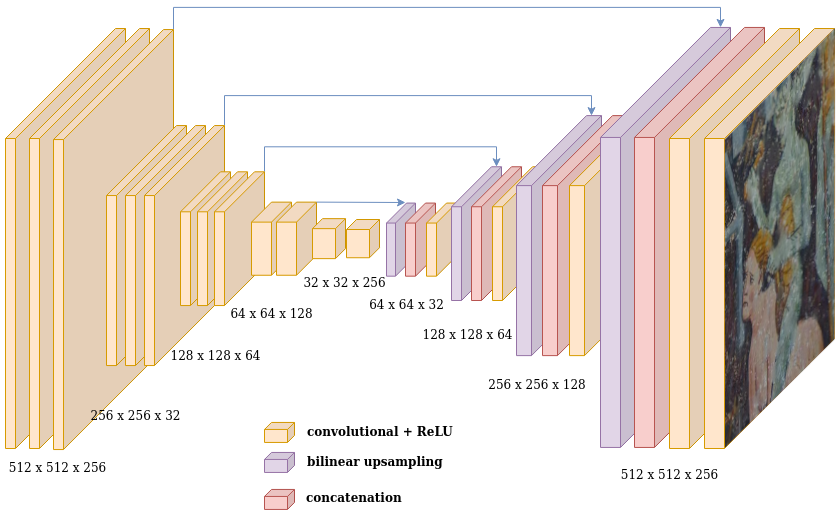}
\label{fig:deep_architecture} 
\caption{The architecture of the DIP network: "hourglass" architecture, downsampling via convolution and upsampling via bilinear upsampling and skip connections.}
\end{figure}

The DIP reconstruction procedure depicted in Figure \ref{fig:dip} makes use of the network architecture represented in Figure \ref{fig:deep_architecture}.
The "hourglass" structure consists of convolutional downsampling and bilinear upsampling with a filter stride equal to 2, whereas the non-linearity considered is a LeakyReLU. 
In more detail, downsampling is achieved via strides and convolution or via max pooling and downsampling with Lanczos kernel. For the upsampling, the two most common approaches are bilinear upsampling and nearest neighbours upsampling. 
Regarding convolutional filters, we tested both filters with the same size and a progressively increasing number for both the encoder and decoder. The size of the filters defines the sensitivity of the convoluted network to different scales of features.
In our experiments, we kept the filter size at 3x3 for all the convolutional layers and we finally chose the reflection padding for more local coherent results in the corner areas. \\
Input and output images are of the same size, i.e. $512 \times 512$ pixels. 
The input image is generally drawn from a multi-variate uniform noise distribution with values in $[0, 1]$.
The performance of the model is significantly impacted by the selection of the optimiser. After evaluating various options, we ultimately decided to use RMSProp (Root Mean Square Propagation) by PyTorch, which exhibited robustness against artefacts. Optimisation was run for 3000 iterations with a learning rate of size 0.01.

Figure \ref{fig:deep_architecture} shows the DIP architecture employed. We make use of skip connections, which are direct links between different parts of the convoluted network. They make information flow not only within the architectural structure but also outside of it, which allows an alternative gradient back-propagation path. This technique proved to be one of the most effective tools in improving the performance of convoluted networks, see, e.g., \cite{test45, orhan2018skip, jimaging9070133}.
However, skip connections are typically viewed as disadvantageous in DIP, because they tend to allow structures to bypass the network's architecture and it may lead to inconsistencies and smoothing effects, as outlined in \cite{Ulyanov_2020}. In our specific scenario, on the other hand, such smoothing effect contributed positively to the overall consistency of the inpainted image.  In Section \ref{sec:results}, the usage benefits of skip connections will be discussed.

Inspired by previous work \cite{Liu2019,Cascarano2021,8682856}, we stabilised the training procedure \eqref{eq:deep_prior_network} by further adding to the loss functional a TV regularisation term, thus considering:
\begin{equation}
\hat{\Theta}\in\text{argmin}_\Theta ~  \lambda||m \odot (f_{\Theta}(z) - \bar{x})||^2 + TV(f_{\Theta}(z)).
\label{eq:deep_prior_network_TV}
\end{equation}
In comparison to \eqref{eq:deep_prior_network}, training under \eqref{eq:deep_prior_network_TV} reduces the sensitivity to the stopping time as the presence of TV (suitably balanced with the data term by $\lambda$) prevents noise overfitting.

\section{Experimental setup}\label{sec:experSetup}


The proposed inpainting workflow consists of three distinct steps. First, given an RGB to inpaint, we perform a basic pre-processing (i.e., resizing) to give it as an input to the DIP model, see Section \ref{ssec:preprocessing}. Next, a masking operator identifying the region to inpaint has to be defined, see Section \ref{ssec:mask_detection}. Lastly, both the input and the mask images are given as an input to the the DIP network whose weights are then optimised to produce the desired inpainting result.


\subsection{Image pre-processing}  \label{ssec:preprocessing}

The RGB images in the available dataset have different resolutions and have different quality. Some of them were taken for documentation purposes and are, generally, low quality. On the other hand, some were taken with high-resolution cameras for the visualisation of fine details. This makes the image dataset not homogeneous, which could be indeed a complication as the architecture neural networks for image reconstruction is typically fine-tuned typically for inputs of specific size and quality.

As discussed below in Section \ref{ssec:networks}, the neural network considered in this work runs on square images 
, for which reason we chose a common image size of $512\times 512$ pixels and used these rescaled data for inpainting. Note that the DIP approach considered requires indeed the whole occluded image as an input. The use of the proposed approach on (overlapping) image patches was therefore not considered in this work but could represent indeed an interesting direction of future research.


\subsection{Mask detection}  \label{ssec:mask_detection}

\begin{figure}
\centering
\includegraphics[width=0.9\textwidth]{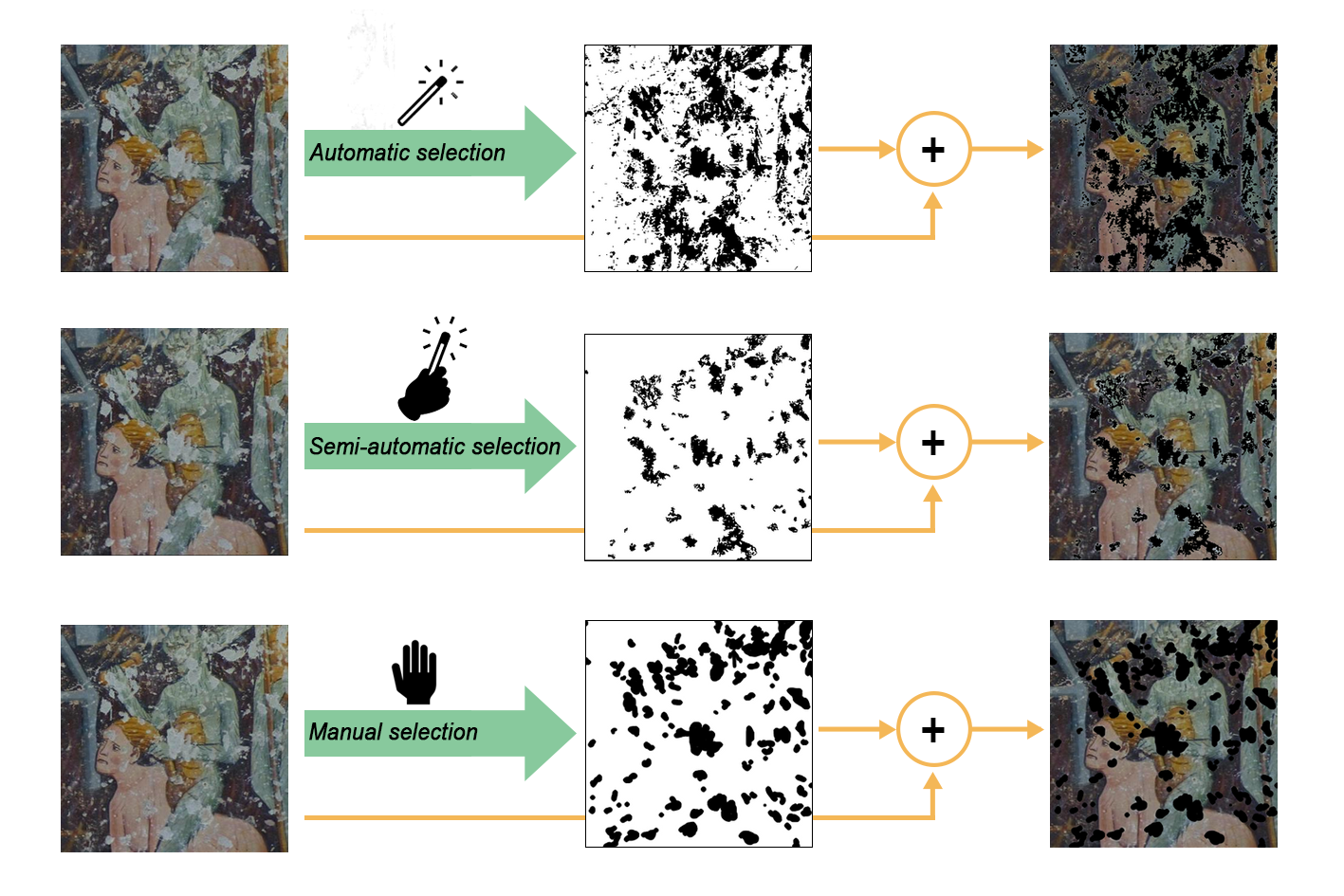}
\caption{Comparison of mask-making methods, for our application the manual method proved to be the most practical.}
\label{fig:workflow_maschere} 
\end{figure}

Computing the pixels in the input image that have to be inpainted is nothing but a binary image segmentation problem which can be handled separately by means of any available segmentation routine.
Such procedure can be approached in different ways, depending on both how much automation one aims to implement and on how relevant the intervention of the restoration professional is. We describe in the following sections three techniques for mask detection falling into the category of automatic, semi-automatic and manual approaches. We stress that other approaches (based, e.g., on the use of deep learning based routines) could alternatively be used.

\begin{figure}
\centering
\includegraphics[width=0.9\textwidth]{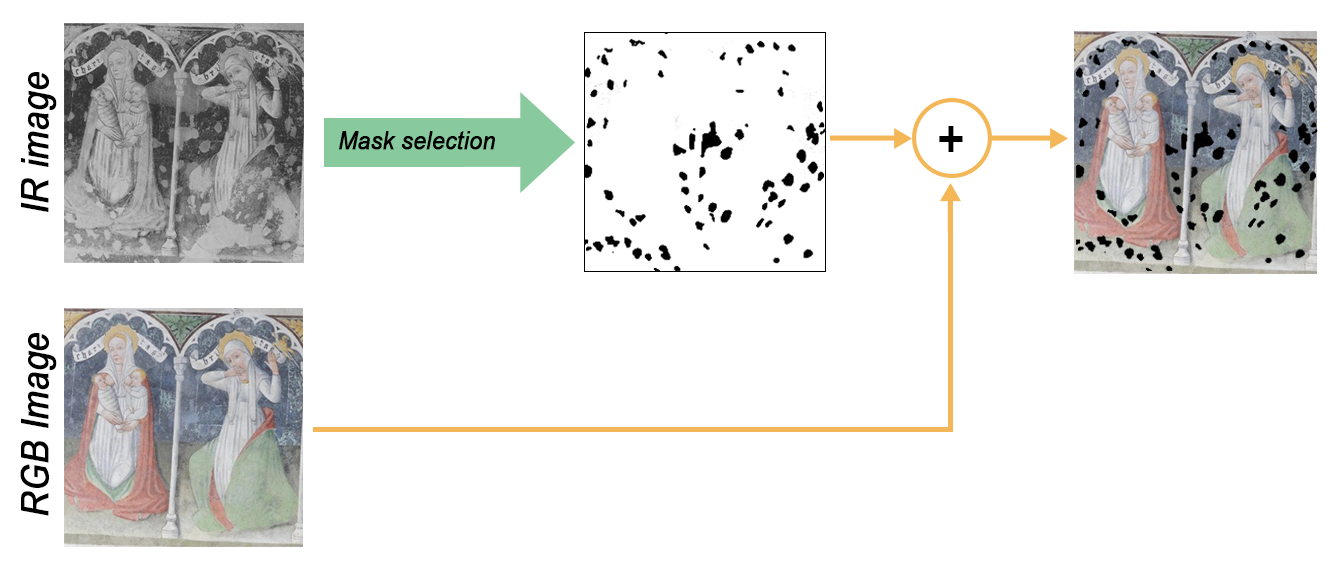}
\caption{ Mask making via an IR version of the RGB image, exploiting IR-enhanced contrasts to effectively select damaged areas.}
\label{fig:workflow_maschere_IR}
\end{figure}

For several RGB images in the PA'INT dataset under consideration, an effective segmentation was not possible due to difficulties in detecting the damaged areas. A valid tool to overcome this issue is the use of infrared (IR) imaging data, which is able to uncover \emph{overpaints}, damages and previous restorations. The inpainting procedure can then be implemented either on the RGB image itself or possibly on the IR image, as schematically reported in Figure \ref{fig:workflow_maschere_IR} and discussed in the following section.

\paragraph{Automatic mask selection.}
For automatic mask selection we refer to a method where an algorithm takes as input a color, corresponding to  the tone of the damaged areas,  and  automatically select all the pixels of that colour  (within a defined tolerance) in  the entire image. 
For our results the threshold was defined on the composite of all three colour channels using GIMP \cite{gimp}. Such procedure works effectively if the damaged areas have considerably distinguishable characteristics with respect to the preserved content, and if this property is consistent throughout the image. If that is not the case and/or too much noise is present in the input data, precision may suffer. 

We found that this techniques was not precise enough for our purposes: additional pixels belonging to the undamaged areas were indeed wrongly detected, see, e.g., Figure \ref{fig:workflow_maschere}.

\paragraph{Semi-automatic mask selection.}
To prevent the mask from including  pixels of the selected colour but not belonging to damages areas, we propose the semi-automatic mask creation.
Unlike to the previous approach, it is done not only by providing a colour and a threshold, but also manually selecting one seed pixel for each connected region of the mask. 
Each region of the mask is then automatically detected  by region growing from the selected pixel.  
Differently from the automatic technique, this approach allows for a better localization of large damages, but  the seed selection may become challenging and potentially imprecise for small regions, as visible in Figure \ref{fig:workflow_maschere}.

\paragraph{Manual mask selection.}
The manual mask selection process involves an expert user utilizing a paint tool to select the damaged areas. This technique is highly effective as it ensures complete coverage of the damage and allows for a customized selection.
By employing this method, we can address the problem of not fully covering the border areas and at the same not extending the mask excessively into the preserved image, as it usually happened with the previous selection methods. 
Leaving portions of the edges of the damaged areas outside the mask, produces discontinuities in the restored images, with a detrimental impact on the quality of the inpainting process.
In our experimental setting, it proved to be the most effective approach in generating the highest quality masks. However, manual mask selection may become impractical due to the considerable amount of manual work involved.



\section{Numerical results}  \label{sec:results}

In this Section, we show the results of the proposed DIP inpainting technique on some images from the PA'INT dataset described in Section \ref{sec:dataset}.

We compare the performance of our DIP approach trained using \eqref{eq:deep_prior_network_TV} (DIP-TV), with the baseline approach in \cite{Ulyanov_2020} (DIP). Whenever skip connections are considered we add ``+skip" to the corresponding approach. When we use TV regularization, the parameter $\lambda$ has been heuristically chosen by minimizing the error metrics of by visual inspection.\\
The DIP-TV+skip solver is compared to state-of-art hand-crafted inpainting models. In particular, we considered the TV-regularisation method  \cite{Chan2001}, the diffusive Navier-Stokes approach \cite{Sapiro}, and the patch-based non-local approach  \cite{Newson2014,ipol.2017.189} with patches of different sizes. 
We remark that fully data-driven inpainting approaches cannot be applied here, as they rely on the use of training data (from the same painter, chapel\ldots) that could not be obtained for our case. 
We ran our experiments on a Ryzen 5600G CPU in tandem with an RTX 3060 GPU. Hand-crafted solvers run on CPU, whereas DIP methods operate on the GPU. Execution times range from approximately 1 second for {\it Navier-Stokes} to 32 seconds for the patch-based non-local approach with a 5x5 patch size, and 81 seconds for size 7x7. For complete convergence, the DIP methods take around 11 minutes. The higher-computational costs are justified by a better reconstruction performance.
The code is available on GitHub at \url{https://github.com/fmerizzi/Deep_image_prior_inpainting_of_ancient_frescoes}

\subsection{Validation on synthetic data}

\begin{figure}[h!]
\begin{subfigure}[b]{0.9\textwidth}
  \centering
  \includegraphics[width=\linewidth]{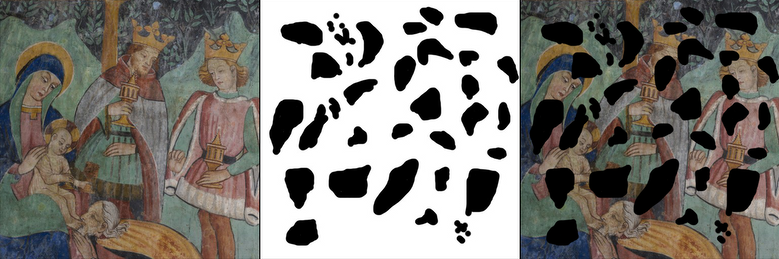}  
   \caption{Original image, Mask and superposition}
  \label{fig:artificial_comparison_setup}
\end{subfigure}
\begin{subfigure}[b]{0.3\textwidth}
  \centering
  \includegraphics[width=\linewidth]{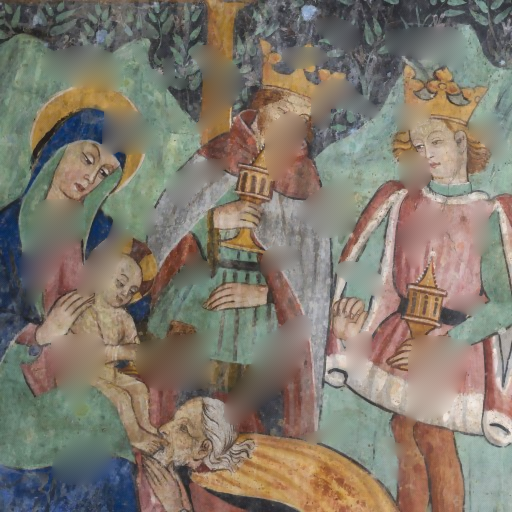}  
  \caption{TV}
  \label{fig:artificial_comparison_patch5}
\end{subfigure}
\begin{subfigure}[b]{0.3\textwidth}
  \centering
  \includegraphics[width=\linewidth]{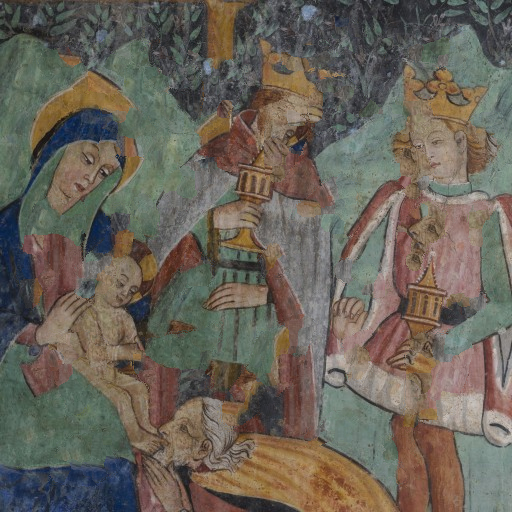}  
  \caption{Patch 7x7}
  \label{fig:artificial_comparison_patch7}
\end{subfigure}
\begin{subfigure}[b]{0.3\textwidth}
  \centering
  \includegraphics[width=\linewidth]{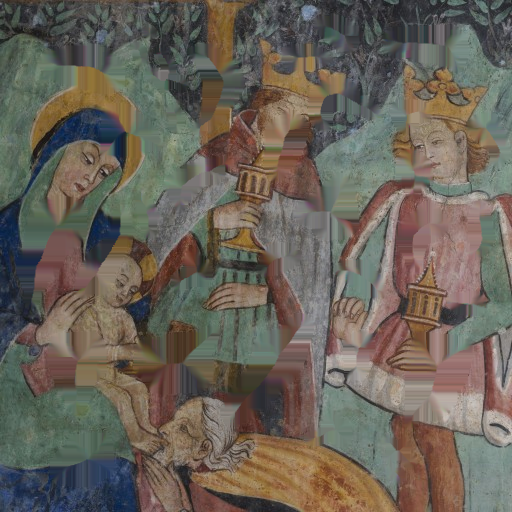}  
  \caption{Navier-Stokes}
  \label{fig:artificial_comparison_navier_stokes}
\end{subfigure}

\begin{subfigure}[b]{0.3\textwidth}
  \centering
  \includegraphics[width=\linewidth]{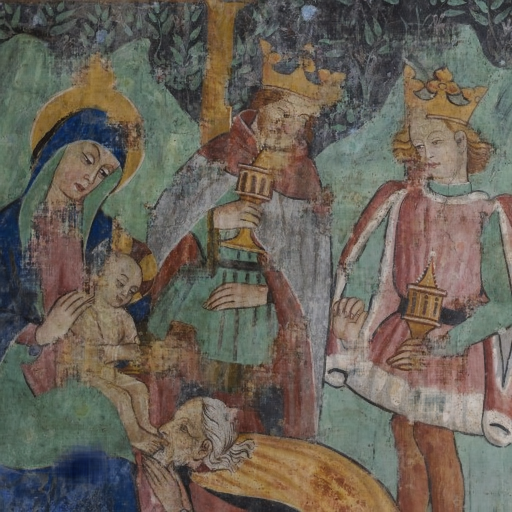}  
  \caption{DIP}
  \label{fig:artificial_comparison_DIP}
\end{subfigure}
\begin{subfigure}[b]{0.3\textwidth}
  \centering
  \includegraphics[width=\linewidth]{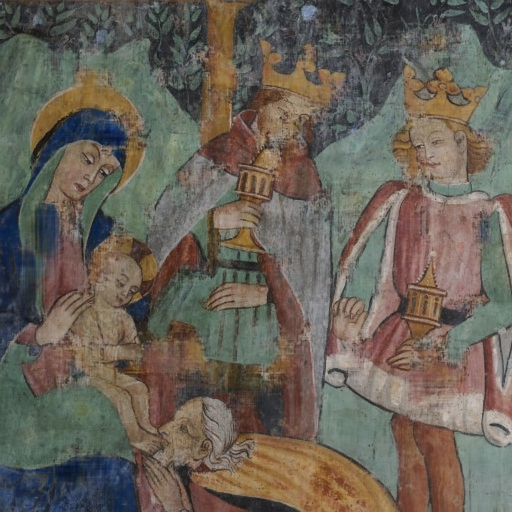}  
  \caption{DIP - TV}
  \label{fig:artificial_comparison_DIP_TV}
\end{subfigure}
\begin{subfigure}[b]{0.3\textwidth}
  \centering
  \includegraphics[width=\linewidth]{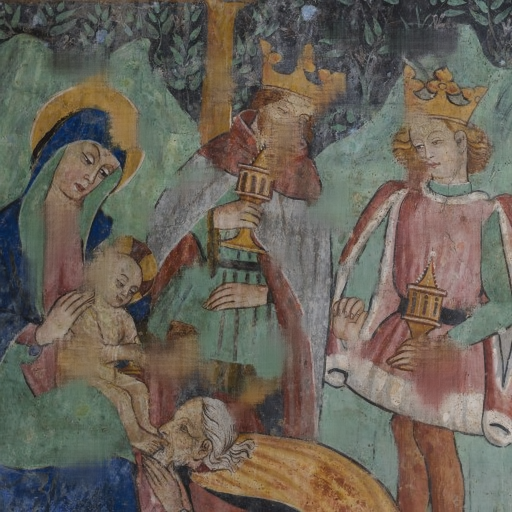}  
  \caption{DIP - TV + skip}
  \label{fig:artificial_comparison_DIP_TV_skip}
\end{subfigure}
\caption{Numerical study simulating the inpainting of an ancient fresco. On the top, the simulation setting with a hand-crafted mask. In the second and third rows, the images inpainted by different techniques, for a visual comparison.}
\label{fig:artificial_comparison}
\end{figure}

We start our numerical discussion presenting some inpainting results obtained from simulated data where an artificially created mask is super-imposed to a representative image in the dataset so to simulate occlusions/damages. We compare the results obtained by hand-crafted approaches and the proposed DIP method and evaluate quantitatively their performance using some standard error measures assessing the quality of the computed reconstruction against the original image.
The original image, the binary mask and the simulated occluded image are reported in Figure \ref{fig:artificial_comparison_setup}. The inpainting results computed using the different methods discussed are reported below. Generally, we observe that the greater the inpainting region, the harder the reconstruction with possibly some non coherent content. \\
We quantitatively assess the reconstruction in terms of the Structural Similarity index (SSIM), the Mean Square Error (MSE), the Normalized Root Mean Square Error (NRMSE) and Peak Signal to Noise Ratio (PSNR).
For all the reconstructions performed, these metrics are presented in Table \ref{tab: artif}.
The computed results consistently highlight that the DIP-TV+skip combination attains the top scores. 

\begin{figure}
\centering
\includegraphics[width=0.9\textwidth]{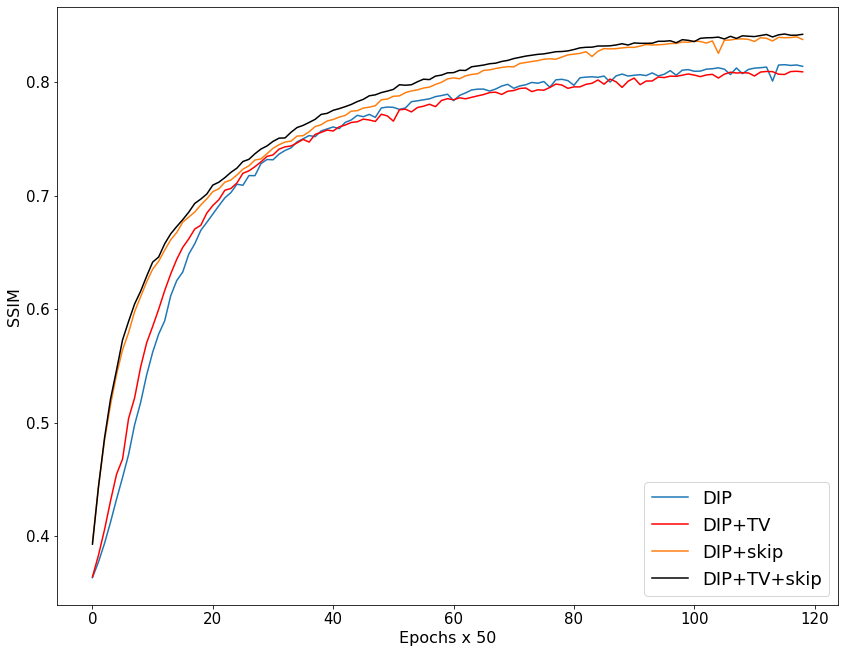}
\caption{Values of the SSIM metric over the training epochs, for four different configurations of the DIP approach.} 

\label{fig:SSIM convergence}
\end{figure}

To highlight the improvement provided by the technical modifications of the DIP scheme detailed in Section \ref{ssec:networks}, in Figure \ref{fig:SSIM convergence} we report the behavior of the SSIM metric over the training epochs, for various DIP configurations.
The naive DIP implementation shows lower SSIM values, in comparison to its versions including skip connections which improve the results throughout all epochs.  We observe that the TV appears to enhance the quantitative results only marginally, although its presence stabilises the training process.
For this reason we considered in the following the DIP-TV+skip combination to perform our tests.

\begin{table*}[htbp]
\centering
\def\arraystretch{1.2}
\begin{tabular}{p{5cm}cccc}
\multicolumn{4}{c}{\textbf{Quantitative Inpainting Assessment on Synthetic Data}} \\
\multicolumn{1}{c}{\textbf{Model}}  & \textbf{SSIM} & \textbf{NRMSE} & \textbf{MSE} & \textbf{PSNR}  \\ 
\textbf{Original Image}   & 1 & 0 & 0 & $\infty$  \\
\textbf{TV}               & 0.82 & 2.44e-01 & 6.24e02 & 20.2     \\
\textbf{Navier-Stokes    }& 0.83 & 1.11e-01 & 1.31e02 & 26.9  \\ 
\textbf{Patch 3x3 }       & 0.75 & 1.65e-01 & 2.89e02 & 23.5  \\ 
\textbf{Patch 5x5 }       & 0.76 & 1.46e-01 & 2.28e02 & 24.5  \\ 
\textbf{Patch 7x7 }       & 0.77 & 1.37e-01 & 2.00e02 & 25.1  \\ 
\textbf{DIP}              & 0.81 & 1.17e-01 & 1.45e02 & 26.5  \\ 
\textbf{DIP - TV       }  & 0.81 & 1.21e-01 & 1.55e02 & 26.2  \\ 
\textbf{DIP - TV + skip}  & \underline{0.84} & \underline{1.04e-01} & \underline{1.15e02} & \underline{27.5}  \\ 

\end{tabular}
\caption{Quantitative assessment of inpainting methods applied to Figure \ref{fig:artificial_comparison_setup}. }
\label{tab: artif}
\end{table*}

\begin{figure}[h!]
\begin{subfigure}[b]{0.9\textwidth}
  \centering
  \includegraphics[width=\linewidth]{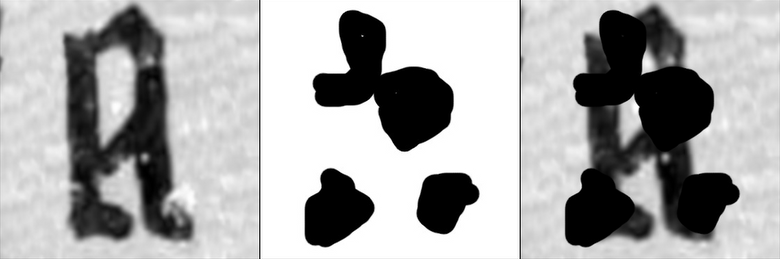}  
  \caption{Original image, mask and superposition}
  \label{fig:letter_setup}
\end{subfigure}
\begin{subfigure}[b]{0.3\textwidth}
  \centering
  \includegraphics[width=\linewidth]{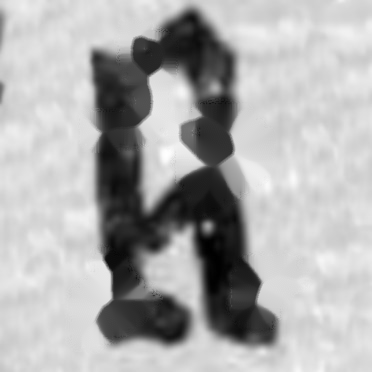}  
  \caption{Navier-Stokes\\SSIM = 8.61e-01}
  \label{fig:letter_NavierStokes}
\end{subfigure}
\begin{subfigure}[b]{0.3\textwidth}
  \centering
  \includegraphics[width=\linewidth]{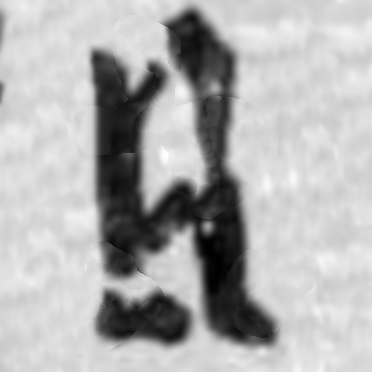}  
  \caption{Patch 5x5\\SSIM = 8.74e-01}
  \label{fig:letter_patch}
\end{subfigure}
\begin{subfigure}[b]{0.3\textwidth}
  \centering
  \includegraphics[width=\linewidth]{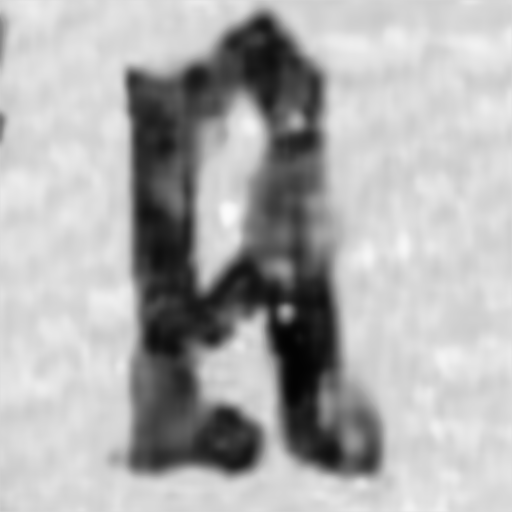}  
  \caption{DIP - TV + skip\\SSIM = 9.03e-01}
  \label{fig:letter_navier_stokes}
\end{subfigure}
\caption{Inpainting of ``a" character with artificial mask}
\label{fig:letter_dip_results}
\end{figure}

We perform a similar simulation on a textual character of an ``a" occluded with an artificially created large inpainting mask, see  Figure \ref{fig:letter_dip_results}.
We compare the solution obtained by DIP-TV+skip with the ones obtained by using the Navier-Stokes and Patch approaches. Both visually and in terms of SSIM we observe that the DIP approach better reconstructs the letter without spots or discontinuities (as in Figures \ref{fig:letter_NavierStokes}-\ref{fig:letter_patch}), showing better visual coherence.

\subsection{Comparison of inpainting techniques on digital pictures of degraded frescoes}

\begin{figure}[h!]
\begin{subfigure}[b]{0.9\textwidth}
  \centering
  \includegraphics[width=\linewidth]{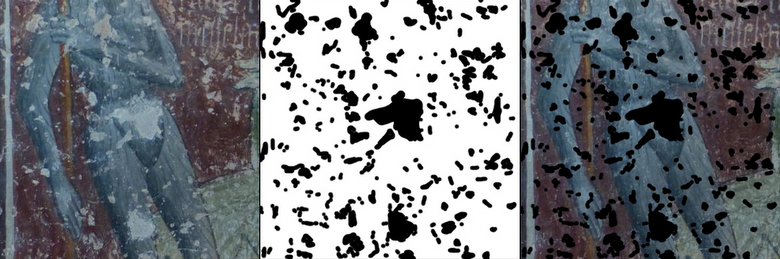}  
   \caption{Original image, Mask and superposition}
  \label{fig:invidia1_setup}
\end{subfigure}
\begin{subfigure}[b]{0.45\textwidth}
  \centering
  \includegraphics[width=\linewidth]{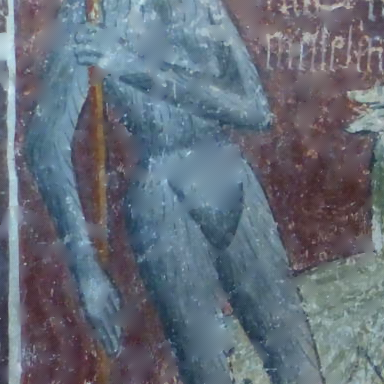}  
  \caption{TV }
  \label{fig:invidia1_TV_inp}
\end{subfigure}
\begin{subfigure}[b]{0.45\textwidth}
  \centering
  \includegraphics[width=\linewidth]{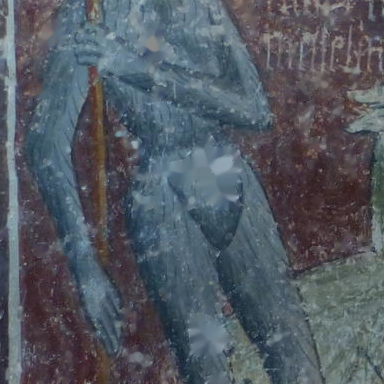}  
  \caption{Navier-Stokes  }
  \label{fig:invidia1_navier_stokes}
\end{subfigure}
\begin{subfigure}[b]{0.45\textwidth}
  \centering
  \includegraphics[width=\linewidth]{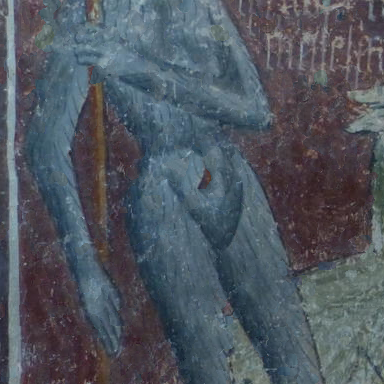}  
  \caption{Patch 5x5}
  \label{fig:invidia1_patch}
\end{subfigure}
\begin{subfigure}[b]{0.45\textwidth}
  \centering
  \includegraphics[width=\linewidth]{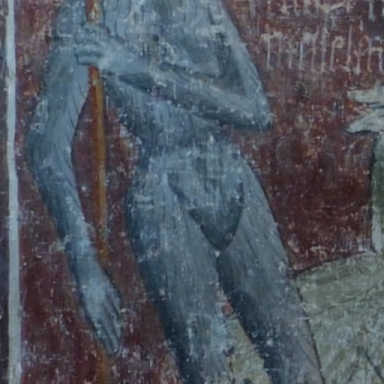}  
  \caption{DIP - TV + skip}
  \label{fig:invidia1_DIP}
\end{subfigure}
\caption{Inpainting comparison on a detail from \emph{Invidia}
}
\label{fig:comparison_invidia1}
\end{figure}

\begin{figure}[h!]
\begin{subfigure}[b]{0.9\textwidth}
  \centering
  \includegraphics[width=\linewidth]{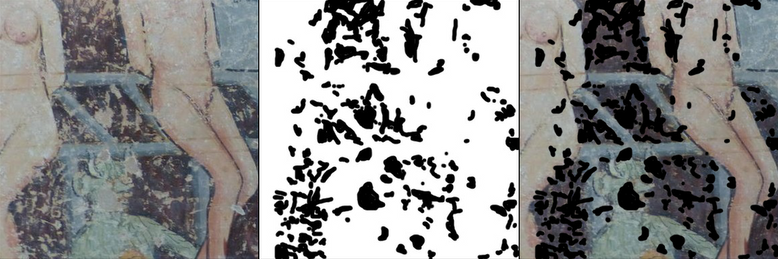}  
   \caption{Original image, Mask and superposition}
  \label{fig:lusuria3_setup}
\end{subfigure}
\begin{subfigure}[b]{0.45\textwidth}
  \centering
  \includegraphics[width=\linewidth]{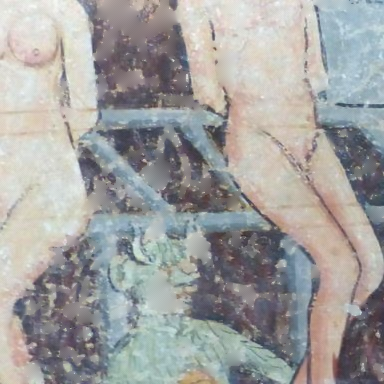}  
  \caption{TV }
  \label{fig:lusuria3_TV_inp}
\end{subfigure}
\begin{subfigure}[b]{0.45\textwidth}
  \centering
  \includegraphics[width=\linewidth]{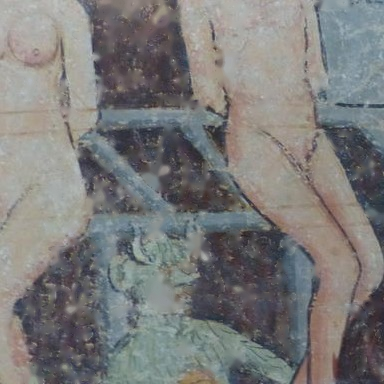}  
  \caption{Navier-Stokes}
  \label{fig:lusuria3_navier_stokes}
\end{subfigure}
\begin{subfigure}[b]{0.45\textwidth}
  \centering
  \includegraphics[width=\linewidth]{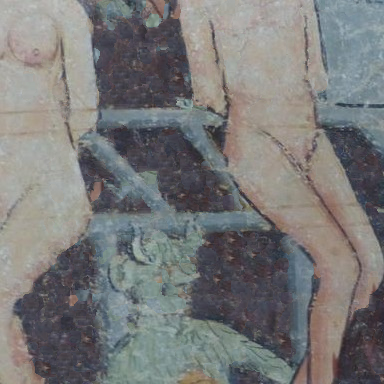}  
  \caption{Patch 5x5}
  \label{fig:lusuria3_patch}
\end{subfigure}
\begin{subfigure}[b]{0.45\textwidth}
  \centering
  \includegraphics[width=\linewidth]{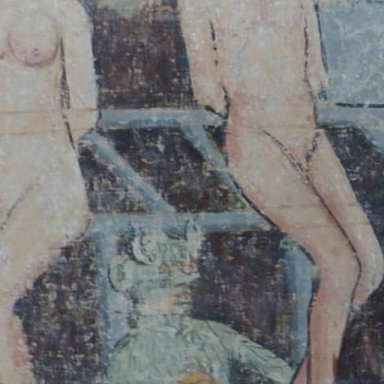}  
  \caption{DIP - TV + skip}
  \label{fig:lusuria3_DIP}
\end{subfigure}
\caption{Inpainting comparison on a detail from \emph{Lusuria}.}
\label{fig:comparison_lusuria3}
\end{figure}

In Figures \ref{fig:comparison_invidia1} and \ref{fig:comparison_lusuria3} we report a comparison between the reconstructions obtained by different inpainting methods tested on the \emph{Invidia} and \emph{Lusuria} frescoes in Figure \ref{fig:san_sebastiano}, respectively. 
 
\begin{figure}[h!]
\begin{subfigure}[b]{0.45\textwidth}
  \centering
  \includegraphics[height=5cm]{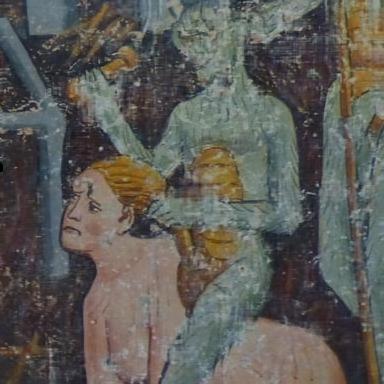}  
  \caption{DIP - TV}
  \label{fig:skipnoskip_NoSkip}
\end{subfigure}
\begin{subfigure}[b]{0.45\textwidth}
  \centering
  \includegraphics[height=5cm]{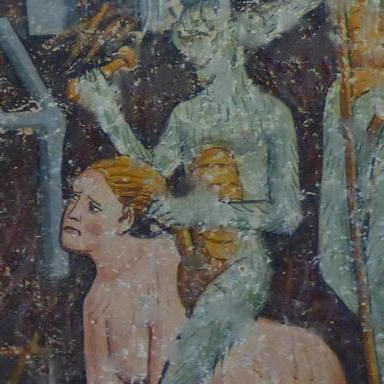}  
  \caption{DIP - TV + skip}
  \label{fig:skipnoskip_WithSkip}
\end{subfigure}
\caption{Comparison of DIP based inpainting without and with skip connections, on a detail from {\it Lusuria}.}
\label{fig:skipnoskip}
\end{figure}

We first consider a cropped image from {\it Invidia}, in Figure \ref{fig:comparison_invidia1}. We note that the TV inpainted image is blurred in the larger damaged regions, whereas the Navier-Stokes image shows evident reconstruction artifacts and the image obtained by the non-local patch-based method is globally better, although a ghosting artifact appears in the largest inpainted area. The DIP-TV+skip inpainting result is the most visually satisfying reconstruction, with fewer artifacts and higher visual consistency. Similar considerations can be made when looking at the results reported in Figure  \ref{fig:comparison_lusuria3}. \\
We remark that the evaluation of results is here only qualitative due to the lack of ground truth images. Recalling reference works in imaging and vision such as \cite{SU1988,gestalt}, the minimal property that should be guaranteed by any inpainting method is the so-called \emph{good connection} property, i.e. the ability of connecting separated pieces of a curve (here, image level lines) in a coherent way. The approaches considered do satisfy this minimal property at least whenever the inpainting domain is sufficiently small. They are subject, however, to more variability in the reconstruction of oscillating content such as, e.g., texture. 
\\
In Figure \ref{fig:skipnoskip}, we present a visual comparison of the inpainting process using DIP, both with and without skip connections. It is evident that incorporating skip connections results in smoother inpainted surfaces and fewer artifacts.

\begin{figure}[h!]
\begin{subfigure}[b]{0.9\textwidth}
  \centering
  \includegraphics[width=\linewidth]{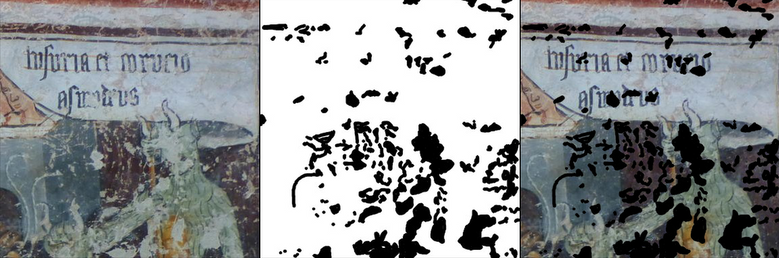}  
  \label{fig:sub-second}
  \vspace{-0.5cm}
    \caption{Original image, mask and superposition}
\end{subfigure}
\begin{subfigure}[b]{0.45\textwidth}
\begin{tikzpicture}[spy using outlines={circle,yellow,magnification=3,size=1.5cm, connect spies}]
\node[anchor=south west,inner sep=0]  at (0,0) {\includegraphics[width=\linewidth]{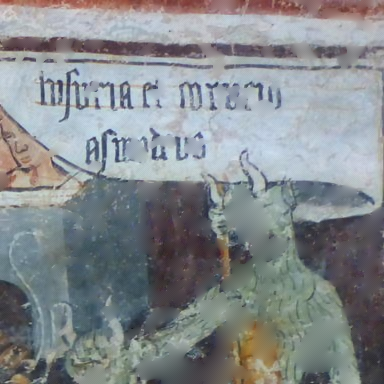}};
\spy on (2.98,3.55) in node [left] at (5.6,0.9);
\end{tikzpicture}
\caption{TV}
\label{fig:DIPonIR_christ_zoom}
\end{subfigure}
\begin{subfigure}[b]{0.45\textwidth}
\begin{tikzpicture}[spy using outlines={circle,yellow,magnification=3,size=1.5cm, connect spies}]
\node[anchor=south west,inner sep=0]  at (0,0) {\includegraphics[width=\linewidth]{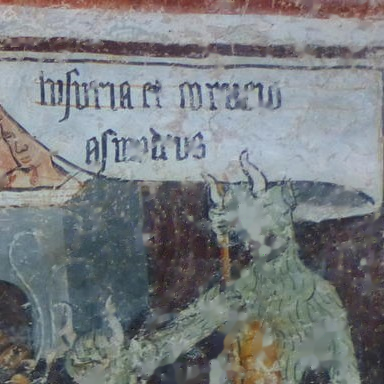}  };
\spy on (2.98,3.55) in node [left] at (1.65,0.9);
\end{tikzpicture}
\caption{Navier-Stokes }
\label{fig:DIPonIR_christ_zoom}
\end{subfigure}
\begin{subfigure}[b]{0.45\textwidth}
\begin{tikzpicture}[spy using outlines={circle,yellow,magnification=3,size=1.5cm, connect spies}]
\node[anchor=south west,inner sep=0]  at (0,0) {\includegraphics[width=\linewidth]{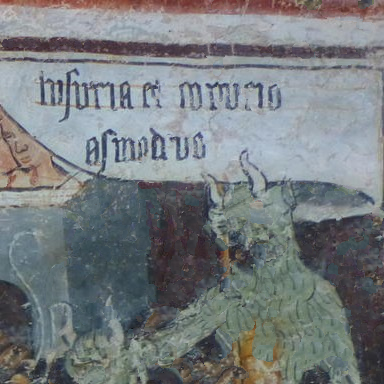}  };
\spy on (2.98,3.55) in node [left] at (5.6,4.8);
\end{tikzpicture}
\caption{Patch 5x5}
\label{fig:DIPonIR_christ_zoom}
\end{subfigure}
\begin{subfigure}[b]{0.45\textwidth}
\begin{tikzpicture}[spy using outlines={circle,yellow,magnification=3,size=1.5cm, connect spies}]
\node[anchor=south west,inner sep=0]  at (0,0) {\includegraphics[width=\linewidth]{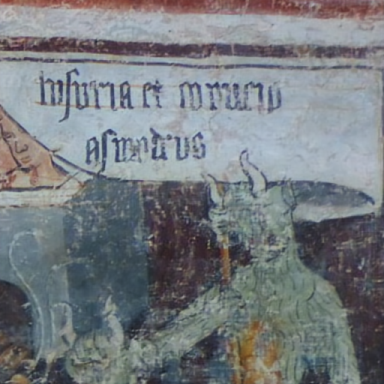}  };
\spy on (2.98,3.55) in node [left] at (1.65,4.8);
\end{tikzpicture}
\caption{DIP - TV + skip}
\label{fig:DIPonIR_christ_zoom}
\end{subfigure}
\caption{Inpainting comparison with a detail of \emph{Lusuria} with both text and figurative parts.}
\label{fig:comparison_lusuria_text}
\end{figure}

\begin{figure}[h!]
\begin{subfigure}[b]{0.85\textwidth}
  \centering
  \includegraphics[width=\linewidth]{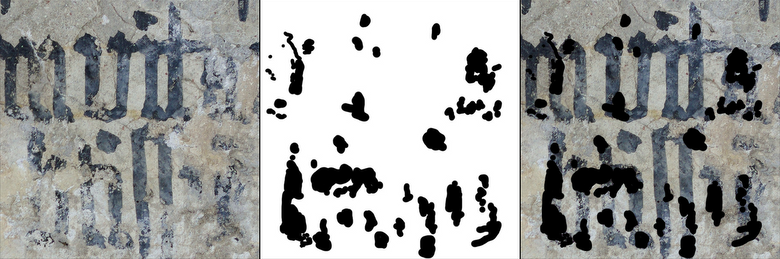}  
  \caption{Original image, mask and superposition}
  \label{fig:venanson_text_setup}
\end{subfigure}
\begin{subfigure}[b]{0.45\textwidth}
  \centering
  \includegraphics[height=5cm]{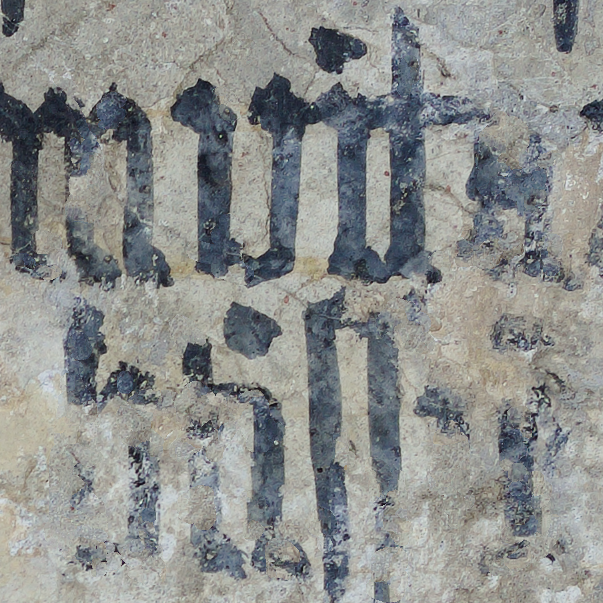}  
  \caption{Patch 5x5}
  \label{fig:venanson_text_patch}
\end{subfigure}
\begin{subfigure}[b]{0.45\textwidth}
  \centering
  \includegraphics[height=5cm]{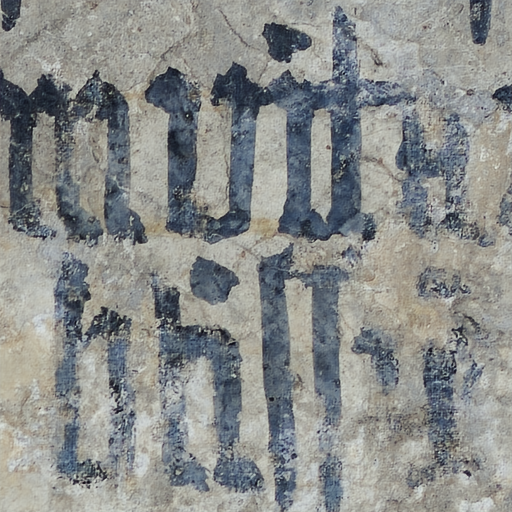}  
  \caption{DIP - TV + skip}
  \label{fig:venanson_text_DIP}
\end{subfigure}
\caption{Text inpainting comparison on a detail from the Venanson chapel.}
\label{fig:venanson_text}

\end{figure}

We now apply inpainting to restore textual images.
The restoration of the textual detail in Figure \ref{fig:comparison_lusuria_text} is particularly interesting. Reliable inpainting approaches should indeed avoid any major modifications to image contents so as to guarantee a reliable, or even improved, interpretation of the artpiece. In this respect, we observe that while local and non-local methods may alter the image content, the DIP approach better preserves the desired text information with a higher level of precision. \\
Analogously, in Figure \ref{fig:venanson_text} we provide a comparison of inpainting methods on a portion of damaged text from the Venanson chapel, where we observe that a more consistent text reconstruction is obtained by our DIP-TV+skip method. 

\subsection{Inpainting based on IR images}   \label{ssec:IR_inp}

\begin{figure}[h!]
\begin{subfigure}[b]{0.45\textwidth}
  \centering
  \includegraphics[width=\linewidth]{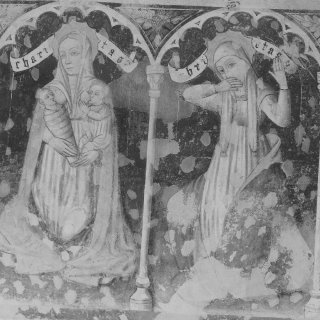}  
  \caption{IR image}
  \label{fig:DIPonIR_christ_a}
\end{subfigure}
\begin{subfigure}[b]{0.45\textwidth}
\begin{tikzpicture}[spy using outlines={circle,yellow,magnification=2.5,size=2.5cm, connect spies}]
\node[anchor=south west,inner sep=0]  at (0,0) {\includegraphics[width=\linewidth]{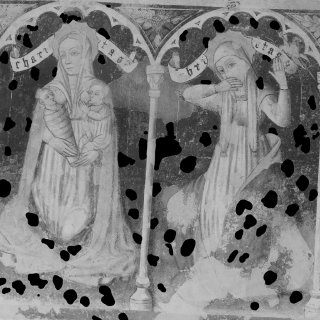}};
\spy on (5.2,2.6) in node [left] at (2.6,1.5);
\end{tikzpicture}
\caption{IR mask}
\label{fig:DIPonIR_christ_b}
\end{subfigure}
\begin{subfigure}[b]{0.45\textwidth}
\begin{tikzpicture}[spy using outlines={circle,yellow,magnification=2.5,size=2.5cm, connect spies}]
\node[anchor=south west,inner sep=0]  at (0,0) {\includegraphics[width=\linewidth]{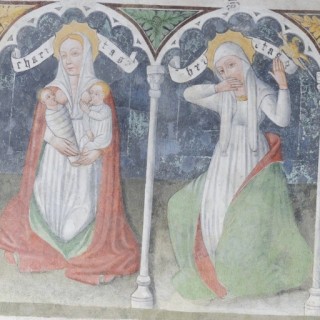}};
\spy on (5.2,2.6) in node [left] at (2.6,1.5);
\end{tikzpicture}
\caption{RGB image}
\label{fig:DIPonIR_christ_c}
\end{subfigure}
\begin{subfigure}[b]{0.45\textwidth}
\begin{tikzpicture}[spy using outlines={circle,yellow,magnification=2.5,size=2.5cm, connect spies}]
\node[anchor=south west,inner sep=0]  at (0,0) {\includegraphics[width=\linewidth]{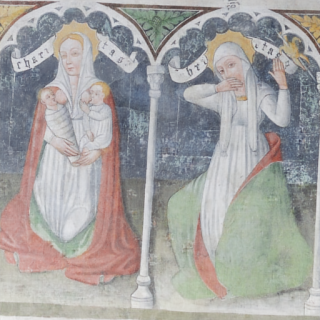}};
\spy on (5.2,2.6) in node [left] at (2.6,1.5);
\end{tikzpicture}
\caption{Inpainted RGB image with IR mask}
\label{fig:DIPonIR_christ_d}
\end{subfigure}
\caption{DIP-TV + skip Inpainting on RGB image with IR mask.}
\label{fig:DIPonIR_christ}
\end{figure}

When an infrared image of a fresco is available, it may allow the discovery of under-drawings and under-writings not easily discernible within the visible spectrum, i.e.~on the RGB image. In Figure \ref{fig:DIPonIR_christ} we exploit such property by creating the mask of these regions using the IR image (Figure \ref{fig:DIPonIR_christ_a}).

Since the  damaged areas are harder to detect (Figure \ref{fig:DIPonIR_christ_c}), the mask has subsequently been super-imposed to the RGB picture of the fresco. DIP  inpainting can there be applied so as to obtain the inpainted image shown in Figure \ref{fig:DIPonIR_christ_d}. In such inpainting result the background looks very coherent to the remaining part of the fresco, thus providing probably a more faithful image of how the original fresco looked like before retouches.

\begin{figure}
\begin{subfigure}[b]{0.45\textwidth}
\begin{tikzpicture}[spy using outlines={circle,yellow,magnification=2,size=2cm, connect spies}]
\node[anchor=south west,inner sep=0]  at (0,0) {\includegraphics[width=\linewidth]{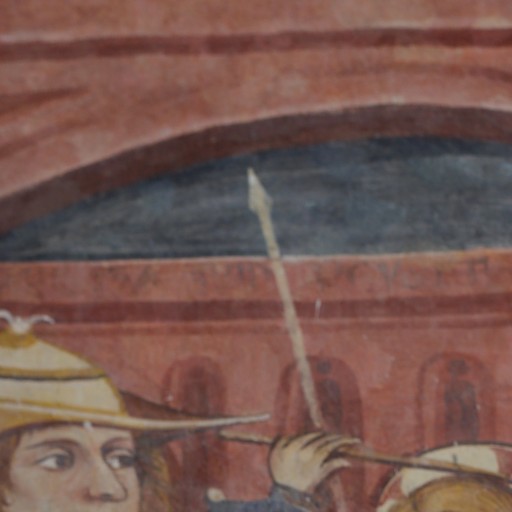}};
\spy on (3.9,2.6) in node [left] at (2.2,4.5);
\end{tikzpicture}
\caption{RGB image}
\label{fig:DIPonIR_arc_a}
\end{subfigure}
\begin{subfigure}[b]{0.45\textwidth}
  \centering
  \includegraphics[width=\linewidth]{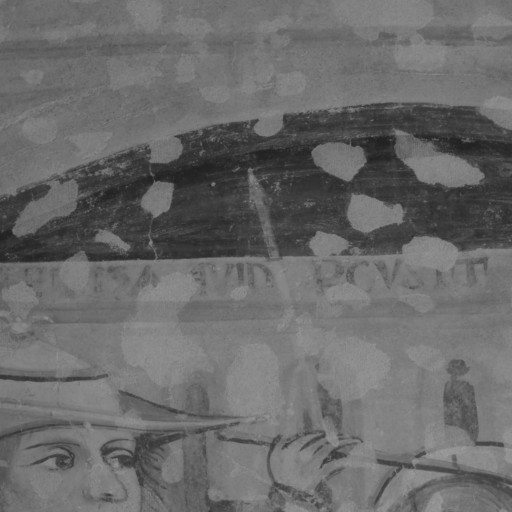}  
  \caption{IR image}
  \label{fig:DIPonIR_arc_b}
\end{subfigure}
\begin{subfigure}[b]{0.45\textwidth}
  \centering
  \includegraphics[width=\linewidth]{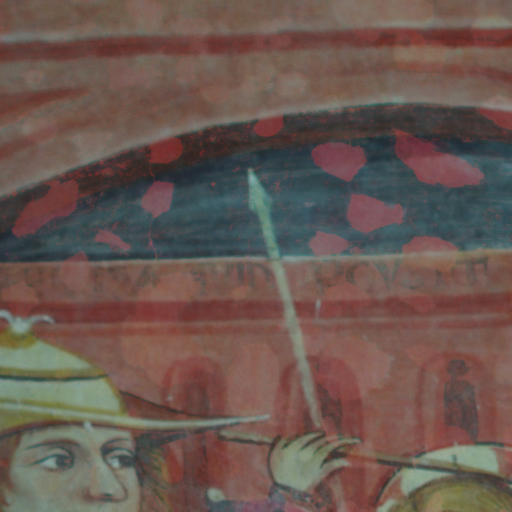}  
  \caption{IR-GB image (before inpainting)}
  \label{fig:DIPonIR_arc_c}
\end{subfigure}
\begin{subfigure}[b]{0.45\textwidth}
\begin{tikzpicture}[spy using outlines={circle,yellow,magnification=2,size=2cm, connect spies}]
\node[anchor=south west,inner sep=0]  at (0,0) {\includegraphics[width=\linewidth]{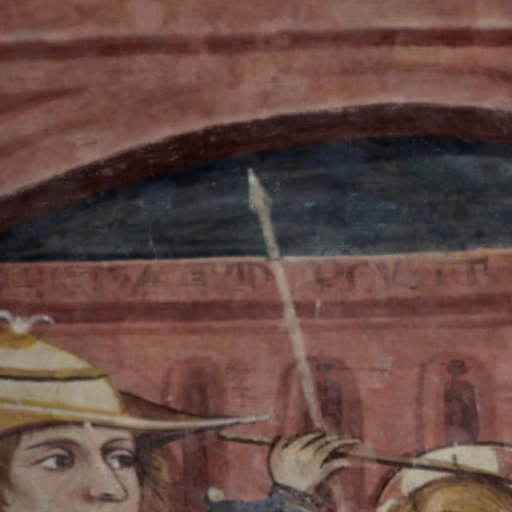}};
\spy on (3.9,2.6) in node [left] at (2.2,4.5);
\end{tikzpicture}
\caption{IR-GB image (after inpainting)}
\label{fig:DIPonIR_arc_d}
\end{subfigure}
\caption{Text enhancing by IR mask extraction. Inpainting is performed by DIP-TV + skip on the IR image. The inpainted IR image is then used as red channel for the original RGB image.}
\label{fig:DIPonIR_arc}
\end{figure}

Interestingly, in the ``Mocking of Christ" painted by Pietro Guido, the IR data revealed ancient text appearing severely faded in the colour image (see  Figures \ref{fig:DIPonIR_arc_a} and \ref{fig:DIPonIR_arc_b}). 
The IR image can be embedded as the Red channel together with the original Green and Blue ones, so as to get the three channel image represented in  \ref{fig:DIPonIR_arc_c} (denoted as IR-GB). 
In this case, the inpainting mask has been selected on the IR picture and used to fill in the IR image directly, by our DIP-TV+skip method. We observe that, now, in the corresponding IR-GB image \ref{fig:DIPonIR_arc_d} the text appears more visible and interpretable than in the starting image \ref{fig:DIPonIR_arc_a}.


\section{Discussion and outlook}  \label{sec:conclusions}

In digital imaging, bringing back to light hidden and/or destroyed piece of information in ancient frescoes using techniques in the realm of variational methods and deep learning is often a very challenging task.  The lack of reference data and the poor quality of both the fresco and of its digital representation often make hopeless the use of both standard approaches based on local reconstruction techniques and complex learning architectures relying on lots of training data. 


In this paper, we consider the problem of image and text inpainting for images acquired in the Mediterranean Alpine arc (dataset PA'INT) and corrupted by severe degradations. The ultimate goal of this project is to ease 
the investigation of the actions taken by the authors toward painted images and their causes, which may emerge in a different context from the period of the artworks' creation.
Intentional destruction and modifications are key aspects we seek to identify in this kind of study. For example, vandalism often targets images with negative connotations, such as devils and demons, leading to the loss of texts and visual representations. The retrieval of these elements is crucial for studying painted themes and patterns which are recurrent during the medieval period .

For such task, we applied the Deep Image Prior Inpainting procedure introduced in \cite{Ulyanov_2020} stabilized as in \cite{8682856} as a hybrid technique relying on the expressivity of (an untrained) neural network and on its interpretability as a non-convex variational approach based on iterative regularisation. By using as a training image the sole given data, improved reconstructions are obtained in the occluded/damaged areas. In comparison with classical approaches, the results computed show less artefacts and favour better interpretability of the data by art historians.

Furthermore, when combined with additional infrared data, the proposed techniques integrate and restore image contents effectively thus providing useful piece of information for subsequent analysis.

Through this interdisciplinary project combining art history, mathematical image processing, and AI, we aim to better understand the historical data and later interventions on medieval images. By doing so, we hope to chronicle the life of the paintings and gain insights into their impact and evolution within past societies.


\begin{backmatter}


\section*{Funding}
PS, FM, OA, RMD and LC acknowledge the financial support received by the CNRS project PRIME Imag'In and the UCA project Arch-AI-story. LC and EM acknowledge the support received by the Academy 1 of UCA, program IDEX JEDI for invited researchers. LC acknowledges the support received by the ANR JCJC project TASKABILE (ANR-22-CE48-0010). Research partially supported by the Future AI Research (FAIR) project of the National Recovery and Resilience Plan (NRRP), Mission 4 Component 2 Investment 1.3 funded from the
European Union - NextGenerationEU.

  

\section*{Availability of data and materials}
The datasets analysed during the current study are available in the PA'INT \cite{paint_web} repository. 
The source code used for DIP inpainting is openly accessible in a dedicated GitHub repository \cite{github}.

\section*{Competing interests}
The authors declare that they have no competing interests.


\section*{Authors' contributions}
OA and PS collected the data. FM developed the computational methods and processed the data. FM, PS, EM and LC analysed the results. PS, RMD provided the historical and artistic background for the project. PS, FM, EM, ELP and LC wrote the manuscript.



\bibliographystyle{bmc-mathphys} 
\bibliography{bmc_article}      

\end{backmatter}
\end{document}